\documentclass[10pt,journal,compsoc]{IEEEtran}

\usepackage{amsfonts}
\usepackage{algorithmic}
\usepackage{algorithm}
\usepackage{array}
\usepackage[caption=false,font=normalsize,labelfont=sf,textfont=sf]{subfig}
\usepackage{textcomp}
\usepackage{stfloats}
\usepackage{url}
\usepackage{verbatim}
\usepackage{graphicx}
\usepackage{cite}
\usepackage{amsmath, dsfont, tabularx, pifont}
\usepackage{multirow, adjustbox}
\usepackage{xcolor,colortbl, booktabs}
\usepackage{amssymb}
\usepackage[capitalize]{cleveref}
\usepackage{float}
\crefname{figure}{Fig.}{Figs.}
\crefname{figure}{Figure}{Figures}
\crefname{section}{Sec.}{Secs.}
\Crefname{section}{Section}{Sections}
\Crefname{table}{Table}{Tables}
\crefname{table}{Tab.}{Tabs.}

\DeclareMathOperator*{\argmax}{arg\,max}

\begin{document}
\setlength{\tabcolsep}{3pt}
\setlength{\abovedisplayskip}{6pt}
\setlength{\belowdisplayskip}{6pt}
\definecolor{Gray}{gray}{0.85}
\newcolumntype{g}{>{\columncolor{Gray}}c}
\newcommand{\cmark}{\ding{51}}
\newcommand{\xmark}{\ding{55}}
\newcommand{\mypar}[1]{\noindent\textbf{#1.}}

 \title{Uncertainty-guided Open-Set Source-Free Unsupervised Domain Adaptation with Target-private Class Segregation}

\author{Mattia Litrico, Davide Talon, Sebastiano Battiato, Alessio {Del Bue}, Mario Valerio Giuffrida, Pietro Morerio
\thanks{Mattia Litrico and Sebastiano Battiato are with University of Catania, Italy.}
\thanks{Davide Talon is with Fondazione Bruno Kessler, Trento, Italy.}
\thanks{Mario Valerio Giuffrida is with University of Nottingham, UK.}
\thanks{Pietro Morerio and Alessio Del Bue are with Pattern Analysis \& 
Computer Vision (PAVIS), Istituto Italiano di Tecnologia (IIT), Genova, Italy.}

\thanks{Primary email contact: \texttt{mattia.litrico@phd.unict.it}.}

\thanks{Manuscript received Month day, Year; revised Month Day, year.}
}
\markboth{Journal of \LaTeX\ Class Files,~Vol.~14, No.~8, August~2021}%
{Shell \MakeLowercase{\textit{et al.}}: A Sample Article Using IEEEtran.cls for IEEE Journals}


\IEEEtitleabstractindextext{%
\begin{abstract}

Standard Unsupervised Domain Adaptation (UDA) aims to transfer knowledge from a labeled source domain to an unlabeled target but usually requires simultaneous access to both source and target data. Moreover, UDA approaches commonly assume that source and target domains share the same labels space. Yet, these two assumptions are hardly satisfied in real-world scenarios. This paper considers the more challenging Source-Free Open-set Domain Adaptation (SF-OSDA) setting, where both assumptions are dropped. We propose a novel approach for SF-OSDA that exploits the granularity of target-private categories by segregating their samples into multiple unknown classes. Starting from an initial clustering-based assignment, our method progressively improves the segregation of target-private samples by refining their pseudo-labels with the guide of an uncertainty-based sample selection module.
Additionally, we propose a novel contrastive loss, named NL-InfoNCELoss, that, integrating negative learning into self-supervised contrastive learning, enhances the model robustness to noisy pseudo-labels. Extensive experiments on benchmark datasets demonstrate the superiority of the proposed method over existing approaches, establishing new state-of-the-art performance. Notably, additional analyses show that our method is able to learn the underlying semantics of novel classes, opening the possibility to perform novel class discovery. 
\end{abstract}

\begin{IEEEkeywords}
Unsupervised Domain Adaptation, Source Free, Open Set, Novel class discovery, Negative Learning, InfoNCE
\end{IEEEkeywords}}

\maketitle

\section{Introduction}
\label{sec:introduction}
Deep learning methods achieve remarkable performance in visual tasks when training and test data share a similar distribution. However, learning approaches struggle to generalise in presence of \emph{domain shift}~\cite{domain_shift,domain_shift2}, namely data coming from different distributions, such as different illumination conditions, imaging modalities or styles. To this end, Domain Adaptation (DA) takes advantage of the knowledge learned on a \emph{source domain} to aid the learning in a different but related \emph{target domain}. As the considerable cost of data collection and annotation limits the availability of large amounts of target data for supervised adaptation of deep learning models, DA focuses on leveraging unsupervised target data for knowledge transfer across domains. In particular, Unsupervised Domain Adaptation (UDA) aims to transfer the knowledge learned on a labelled source domain to an unseen target domain without requiring any target label~\cite{uda1,uda2,uda3,uda4,tpami,tpami2,tpami3}.  

\smallskip
Most UDA techniques typically rely on two strong assumptions on the data. On the one hand, they require simultaneous access to source and target data during the adaptation, hindering the deployment in  applications where data privacy or transmission bandwidth become critical issues. To overcome this issue, recent works focused on the setting of Source-free Domain Adaptation (SF-DA) \cite{Litrico,sfda,sfda1,sfda2,sfda3}, where \textit{source data is no longer accessible} during the adaptation phase, but only unlabelled target data is available. On the other hand, most UDA algorithms assume a \emph{closed-set} setting, \textit{i.e.} the two domains share the same class space. Despite the adaptation to the new domain, the transferred source model has no capability to make predictions for novel/unseen classes. This severely limits their applicability in real-world scenarios \cite{open4,open7}, where the target domain might contain unknown classes that are not present in the source data. 

To address this issue, Open-set Domain Adaptation (OSDA) \cite{open1, open2, open3, open4, open7} aims to empower the adapting model with the ability to recognise novel-class samples in the target domain as \textit{unknown}. In this scenario, the target dataset contains both classes already seen in the source domain (\textit{shared} classes), as well as novel classes that are not present in the source (\textit{private} classes). 
In \cite{open3,open4}, the authors used adversarial learning to align the source domain with the portion of the target domain sharing the same classes, excluding the target private classes, leading to a poor decision boundary. To avoid this issue, Jang et al. \cite{open2} included the target private classes in the adversarial learning to align the source domain with the shared classes, while segregating private samples in the unknown classes. However, all these previous works are not source-free, as they learn from source and target data simultaneously.

The Source-Free Open-set Domain Adaptation (SF-OSDA) addresses both challenges, assuming that the adaptation process has no access to source data in presence of private classes in the target domain. While this joint setting is underexplored, some recent approaches \cite{sfda2,aad,usd} have been proposed. 
However, these approaches consider samples from target-private classes as outliers and cluster them into a \textit{single} ``unknown'' class. 
By gathering in a single class semantically different samples, this yields a degradation of the feature space.
Additionally, to segregate samples of unknown classes, previous approaches rely on a statistic-based separation criterion computed on the outputs of the adapted model, often leading to the so called \textit{negative transfer}, a consequence of the well-known \textit{confirmation bias} \cite{confirmation_bias}.
Namely, the model relies on its noisy outputs for the adaptation process, increasing the confidence in its wrong predictions (in essence, the model is overfitting its own noise).

In this work, we propose a novel Source-free Open-set Domain Adaptation approach that, leveraging the granularity of target-private classes, segregates their samples into \textit{multiple} clusters, resulting in a more effective adaptation compared to previous works. As a by product, this strategy also allows the discovery of the underlying semantics of novel classes.
As an extension of \cite{Litrico}, our method produces pseudo-labels for all target samples and progressively refines them using the consensus from neighbours samples. However, differently from \cite{Litrico}, the initial pseudo-labels assignment (which also includes target private classes) is obtained by clustering the features space spanned by the source model, rather than directly using its predictions. A subset of the obtained clusters centroids is matched with the source model prototypes computed from the classifier weights. The unmatched centroids are treated as new prototypes of the target private classes and used to initialise the columns of the classifier weights corresponding to the unknown classes. Differently from the state-of-the-art, this strategy does not require any separation criterion to distinguish between samples of shared and unknown classes, limiting the negative effects of the confirmation bias.

The pseudo-labels produced during the adaptation are inevitably affected by the so called \textit{shift noise} \cite{morerio2020generative}, namely the noise caused by the domain shift, here further intensified by the presence of novel classes in the target. Moreover, the initial clustering-based assignment could also introduce some noise, potentially leading to classify an entire shared class as unknown or viceversa. To obtain robustness to such noise, we perform an uncertainty-based samples selection to discard sample with unreliable (and possibly noisy) pseudo-labels. To that end, we propose a novel uncertainty estimation solution, which extends the one used in \cite{Litrico} that was not devised to handle this scenario. The uncertainty of the pseudo-labels  is measured by analysing the consensus of neighbours samples predictions, and the relative distances of samples with respect to the class prototypes. Such newly introduced strategy assigns low uncertainty values when samples are close to one prototype, while being far from the others. Contrarily, when the distance values are similar, samples will have a high uncertainty and their pseudo-labels are considered unreliable.

To obtain a regularised features space where neighbours samples are semantically similar, we leverage a self-supervised contrastive learning framework together with a negative pairs exclusion strategy. Differently from \cite{Litrico}, we introduce the \textit{NL-InfoNCELoss}, a novel contrastive loss that integrates the principles of negative learning in the standard contrastive loss \cite{contrastive_testtime}. By empowering the standard loss with the well known benefits of the negative learning paradigm \cite{JPNL,nel}, this formulation increases the robustness of the contrastive framework to the noise inevitably affecting the pseudo-labels.

Extensive experiments demonstrate the effectiveness of our method in the challenging setting of SF-OSDA, outperforming the state-of-the-art on two major benchmark datasets. Specifically, on Office31 \cite{office31} and Office-Home \cite{officehome}, we set the new state-of-the-art by improving the performance of $+2.1$ and $+0.1$ $\text{HOS}$, respectively. Ablation studies show the effectiveness of the new components in handling the issues introduced by the presence of novel classes in the target domain. Notably, additional analyses show the ability of our model to classify samples from target-private classes, meaning the model learned the underlying semantics of novel classes without being explicitly optimised for that task. 

In summary, the main contributions of this work are the following:
\begin{itemize}
    \item We tackle the Source-Free Open-Set UDA problem by exploiting the granularity of target-private classes via an uncertainty-guided pseudo-labelling. As a byproduct, we explicitly learn the underlying semantics of target-private classes, unveiling the possibility of performing novel class discovery;
    \item We introduce \textit{NL-InfoNCELoss}, a novel contrastive loss that integrates the negative learning paradigm into self-supervised contrastive learning to regularise the feature space, while increasing the robustness to  pseudo-label noise;
    \item We evaluate the proposed methodology with extensive experiments proving the superiority against existing approaches on two major benchmark datasets.
\end{itemize}
    
The rest of this paper is organised as follows: \Cref{sec:relatedwork} introduces and discusses the relevant literature related to Open-Set domain adaptation, pseudo-labelling, and learning from noisy labels. \Cref{sec:method} formalises the Open-Set Source-Free UDA setting and presents the proposed methodology. \Cref{sec:experiments_results} presents experiments and additional analyses to assess our approach. Eventually,  \Cref{sec:conclusion} concludes the paper with a summary and a discussion of major findings.

\section{Related Work}
\label{sec:relatedwork}

    \mypar{Open-set and Source-free UDA}
    The assumption that both source and target domains share the same label space may not hold in several real-world scenarios, where the target domain can contain classes that are not included in the source domain. In the literature, such classes are referred to as \textit{target private} (or simply \textit{private}) or \textit{unknown}. Open-set domain adaptation aims to transfer knowledge from a source domain to a target domain containing private classes. Recent works mainly focus on separating the known and unknown samples in the target domain \cite{open1, open2, open3, open4} or aligning the known distributions \cite{open3, open5}. Differently, \cite{open6} proposed to use word-prototypes based on mid-level features, which are more robust to the negative transfer to perform a subsidiary prototype-space alignment.  UADAL \cite{open2} proposed to segregate the unknown classes leveraging a 3-way adversarial learning among the source, the shared, and unknown classes of the target. However, all these methods assume the availability of source data during the adaptation, \textit{i.e.} they are not source-free.
    
    More recently, \cite{onering, generative2, FS} proposed source-free strategies for the open-set domain adaptation setting. However, these approaches rely on ad-hoc pre-training strategies on the source domain that assume the awareness of the presence of unknown classes in the target during the pretraining.
    For example, OneRing \cite{onering}  pretrained a classifier on the source data with an extra class to empower the model to better segregate target-private samples during adaptation. In  \cite{generative2,FS}, the authors leverage a weighting scheme and feature-splicing to avoid negative transfer, but they generate extra training samples of unknown categories during pretraining on the source domain, facilitating the separation of private classes in the target domain. Differently, our method makes no ad-hoc assumptions on the training in the source domain, providing a more general approach to this challenging setting, widening its application.
    
    Other methods have been proposed to avoid such ad-hoc training strategies. SHOT \cite{sfda2} and AaD \cite{aad} use a separation criterion based on the entropy of predictions, to distinguish between shared and unknown classes. Instead, USD \cite{usd} proposed to segregate private samples based on Jenson-Shannon divergence. However, since these separation criteria are computed on model outputs, they are easily affected by confirmation bias. Differently, we do not use any explicit separation criterion, by relying on a clustering-based strategy to initially assign samples to classes, either shared or unknown. Assuming that such assigned labels are noisy, we refine the assignment via pseudo-labelling and self-supervision strategies. 
    
    \mypar{Self-supervised Learning and Pseudo-Labelling}
    Self-supervised methods are successful in learning transferable representations of visual data \cite{self_supervised1, cl1, self_supervised2, cl2, self_supervised3, cl3, moco, self_supervised4, self_supervised5}. Specifically, \cite{cl1,cl2,cl3} use contrastive-based pretext tasks to enhance the generalisation ability of deep models. Moreover, several recent self-supervised approaches have been utilised within UDA \cite{self_supervised6, self_supervised7} and SF-UDA \cite{sfda,ontarget,contrastive_testtime, maracani2024key} settings. On the other hand, pseudo-labelling is a simple but effective technique used in semi-supervised \cite{pseudolabel_1, fixmatch} and self-supervised learning \cite{self_supervised1}, as well as in domain adaptation \cite{sfda2, pseudolabel_2, ontarget, contrastive_testtime}. It consists in using labels predicted by the model as self-supervision. Fix-Match \cite{fixmatch} and On-target \cite{ontarget} take advantage of pseudo-labels without any refinement during training. In this paper, we perform refinement steps to progressively improve the quality of the generated pseudo-labels.
    
    \mypar{Noisy Labels}
    Noisy training sets often result in poorly trained models. In \cite{noise_overfitting}, the authors demonstrated that a deep neural network can easily overfit an entire dataset with any ratio of corrupted labels, resulting in a poor generalisation on test data. To address this problem, different approaches have been proposed focusing on noise-robust losses \cite{noise_robust_loss, MAEGhosh, NCE, NLNL}, estimation of the noise-transition matrix \cite{noise_transition_matrix}, reweighting of the loss based on the reliability of samples \cite{reweighting, reweighting3, reweighting4}, as well as the selection of clean data from noisy samples \cite{sample_selection, DivideMix}. In \cite{sample_selection,DivideMix}, the authors proposed sample selection strategies based on the loss, but they require multiple deep neural networks to make the selection more robust, leading to an increase of the computational cost of the method. Furthermore, the sample selection is based on the loss, which is computed with noisy labels that are not refined during the training. This implies that the amount of noise in the labels will not be reduced and the selection will be highly affected by the noise.
    Instead, we propose a sample selection strategy based on the uncertainty (reliability) of pseudo-labels, which is estimated after their refinement using the consensus among neighbours, without requiring the use of multiple deep networks. Differently from \cite{Litrico}, which used a reweighting strategy on target samples, we perform a hard sample selection to account for the increased amount of noise produced by the unknown classes.
    Lastly, NEL \cite{nel} combined a \textit{negative learning} loss with a pseudo-labels refinement framework based on ensembling. Negative learning  \cite{NLNL} refers to an indirect learning method which uses complementary labels to combat noise. While we also use a negative learning loss, we do not require an ensemble of networks to perform refinement, which results in a reduced computational cost.
    Additionally, we leverage the paradigm of negative learning to also propose a novel contrastive loss empowered with more robustness to the noise.
    
    \mypar{Summary} This paper extends our previous work \cite{Litrico}, which proposed a methodology for source-free domain adaptation in a closed-set scenario. Differently than \cite{Litrico}, here we focus on the open-set setting, where the target domain contains novel classes that are not present in the source.
    As in \cite{Litrico}, we train the target model with generated pseudo-labels that are progressively refined during the adaptation. However, the initial pseudo-labels assignment is obtained by performing a clustering on the target feature space, instead of directly leveraging the source model. Moreover, we propose a novel uncertainty measure devised to limit the impact of the additional noise in the pseudo-labels produced by the presence of unknown classes in the target domain. In \cite{Litrico}, we reweighted the contribution of the pseudo-labels in the loss based on their uncertainty. Here we introduce an uncertainty measure that is combined with the one proposed in \cite{Litrico}, to select (instead of reweight) samples with reliable pseudo-labels, which improves the robustness to the noise in both shared and private classes. Finally, we propose a novel contrastive loss that incorporates the principles of negative learning in the contrastive framework of \cite{Litrico}, to further improve the robustness of the model against the noise.

\section{Proposed Method}
\label{sec:method}

\begin{figure*}[ht]
    \centering
        \includegraphics[width=1.\textwidth]{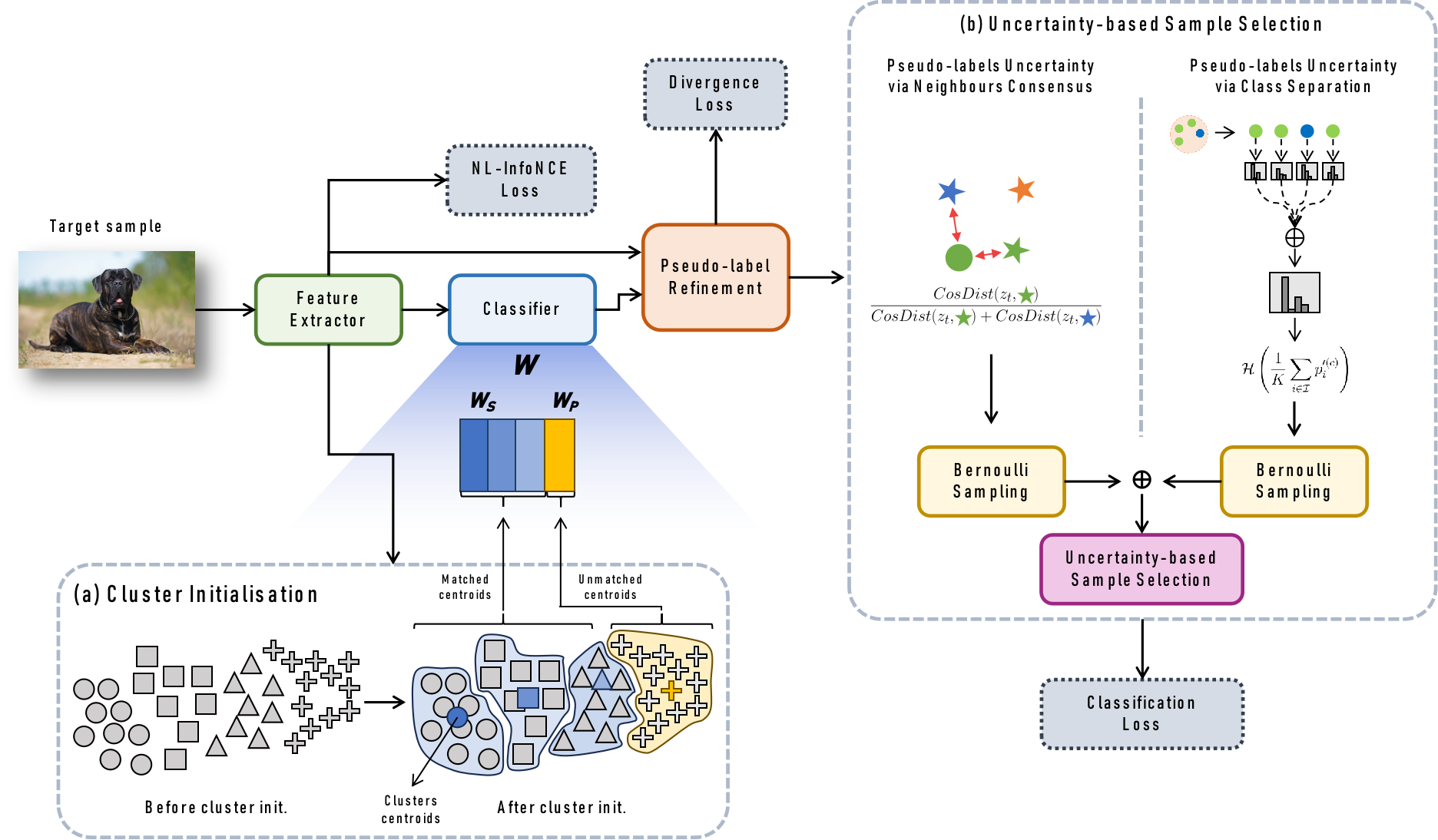}
        \caption{\label{fig:concept}Overview of the proposed adaptation approach. \textbf{(a)} Target samples are clustered based on the features extracted from the pretrained source model, to provide an initial pseudo-labels assignment. Next, clustering allows the identification of class centroids (coloured shapes), which are matched against the most similar prototypes of shared classes. Such prototypes are taken as the columns of weight matrix $W$ of the classifier ({\textsc{blue}} clusters). Since the target domain has more classes than the source domain, some clusters are left out from this matching, which will be treated as target-private classes ({\textsc{yellow}} clusters). \textbf{(b)} After the refining of the pseudo-label, its uncertainty is estimated using two approaches. \textit{Neighbours consensus:} the uncertainty is determined by analysing the consensus of neighbours on the refined pseudo-label. \textit{Class separation:} the model uncertainty is estimated based on the distance w.r.t. the two closest class prototypes. A novel contrastive loss (NL-InfoNCELoss) gathers same-class samples in the features space, while being robust to noisy pseudo-labels.}
        \label{fig:method}
    \end{figure*}
    
  This section presents our approach to the SF-OSDA problem in classification tasks, which is depicted in \Cref{fig:method}.
  Our model is firstly trained on a source domain without using any ad-hoc training strategies, as done in \cite{onering, generative2, FS}. Then, at the beginning of the adaptation process, the pretrained source model extracts the features for each of the unlabelled target samples. A clustering-based strategy on target features roughly detects samples from private classes and provides an initial pseudo-labels assignment on the target samples (\Cref{subsec:initial_clustering}). Due to the domain shift between source and target domains and the presence of unknown class samples, this initial assignment is unreliable. Therefore, we mitigate the impact of the noise in the pseudo-labels by refining them and selecting reliable samples for the training. On the one hand, the nearest neighbors voting scheme in \Cref{subsec:nnvoting} progressively refines the pseudo-labels according to a majority rule. On the other, in \Cref{subsec:entropy_estimation} we select reliable pseudo-labels with low uncertainty, which are then used in the adaptation process.  The contrastive scheme proposed in \Cref{subsec:temporal_queue_exclusion} enforces the underlying assumption of the pseudo-labels refinement process: samples of the same class should be close in the target feature space. 
  In \Cref{subsec:adapt-pseudlabel}, we leverage the pseudo-labels to adapt the model to the target domain through a negative learning classification loss. In \Cref{subsec:overallframework}, we provide the overall training objective that progressively refines the noisy pseudo-labels and segregates target-private samples, while aligning the pretrained model to the target domain.
\smallskip

\mypar{Problem formulation} Let $\mathcal{D}_s$ be the labelled source dataset including pairs $\{x_s, y_s\} $, where $x_s \in \mathcal{X}_s$ and $y_s \in \mathcal{Y}_s$ are images and ground-truth labels, respectively. Due to the source-free constraint, the source data $\mathcal{D}_s$ is not available during adaptation.
Similarly, let $\mathcal{D}_t$ be the target data composed of images $\{x_t\}$ only, with  $x_t \in \mathcal{X}_t$. The open-set scenario assumes that $\mathcal{Y}_s \subset \mathcal{Y}_t$, where $C_S := \mathcal{Y}_s \cap \mathcal{Y}_t$ and $C_P := \mathcal{Y}_t \setminus \mathcal{Y}_s$ are the set of shared and private classes, respectively. The set of all classes $C$ is given by $C=C_S \cup C_P$.
    The source model $g_s (\cdot) = h_s(\varphi_s(\cdot))$ is composed of a features extractor $\varphi_s: \mathcal{X} \rightarrow \mathbb{R}^D$ and a classifier $h_s: \mathbb{R}^D \rightarrow \mathbb{R}^{|C_S|} $, where $D$ is the size of the features space. 
    Without loss of generality, we assume that the weights of the classifier are characterised by a matrix $W \in \mathbb{R}^{D\times |C_S|}$. We will refer to each column $i$ of the weight matrix $W[i]$ as a shared class \textit{prototype}. 
    


    \label{sec:overview}

    \subsection{Leveraging the granularity of target-private classes}
    
    Open-set source-free domain adaptation considers a two-steps approach. In the first stage, a model is trained on large labelled source data. Without further access to the source samples, the second step adapts the source model to the unlabelled target domain. During the (pre-)training on the source dataset, the classifier $h_s$ is trained to accommodate the classes in $C_S$, \textit{i.e.} the shared classes. The target domain contains not only the classes in $C_S$, but also includes a set of private classes $C_P$. 

    Previous works usually train models to segregate samples from target-private classes into a single ``unknown'' class, which is more affine to outlier-detection. While samples from shared classes are separated with respect to their semantics, this strategy forces the model to aggregate samples from target-private classes into a single class, independently from their semantics. This affects the learned feature space, leading to a suboptimal geometry, where samples with possibly very different semantics are grouped together. Moreover, this strategy hinders the possibility to discover the underlying semantics of novel classes.
    To solve this issue, we design our model to leverage the granularity of target-private classes by segregating their samples into multiple unknown classes. This choice has the side effect of learning a features space where samples from private classes are aggregated based on their semantics, as discussed in \Cref{subsec:analysis}.
    Hence, before the adaptation process, we craft a new classifier $h_t$ whose weight matrix is made of two parts, such that $W_T = [W_S|W_P]$. The set of weights $W_S$ is initialised by taking the pretrained weights from $h_s$, whereas $W_P$ is randomly initialised with values sampled from a uniform distribution. This means that we augment the pretrained classifier by adding a certain number of columns matching a plausible number of private classes $|\widehat{C}_P|$ in the target domain. 
    Note that $|\widehat{C}_P|$ is arbitrarily chosen, since the real cardinality of the target private classes is unavailable under the open-set setting. In \Cref{subsec:analysis}, we analyse the performance with different values for $|\widehat{C}_P|$.
    Using the extended classifier $h_t$, we use the model  $h_t(\varphi_s(\cdot))$ for an initial pseudo-label prediction in the target dataset $\mathcal{X}_t$ (\Cref{fig:cluster_init}(a)).  

    \subsection{Clustering-based Target Model Initialisation}
    \label{subsec:initial_clustering}

\begin{figure*}[ht]
        \centering
        \includegraphics[width=\linewidth]{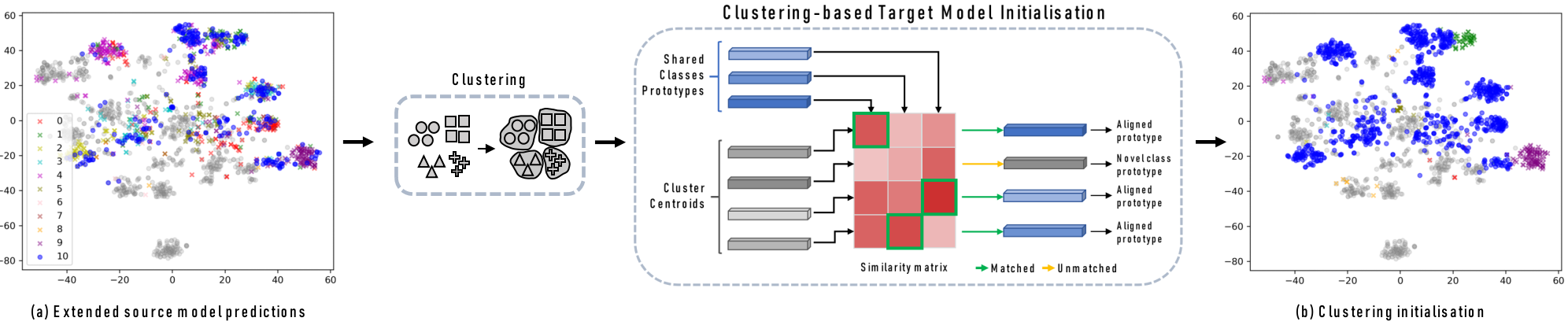}
        \caption{The extended source model $h_t$ incorrectly classifies samples from private classes as belonging to shared ones. Nonetheless, samples from both shared and private classes exhibit a low intra-class variability. Building on this observation, we cluster the feature space to perform an initial pseudo-labels assignment on the target domain. The figure shows the initial identification of target-private samples using clustering. In (a) and (b) coloured samples belong to target-private classes, while \textsc{gray} samples belongs to shared classes. The colour of samples represents the predicted class. For visualisation purposes, all samples predicted in a target-private class are represented in the \textsc{blue} class. The $\bullet$ symbol represents correct predictions, whereas the $\times$ represents incorrect predictions. While the source model cannot correctly assign target-private samples, the clustering leverages the structure in the features space to aggregate private samples and provides a good pseudo-labels initialisation. In fact, after the clustering, more target-private samples are correctly classified in the \textsc{blue} class. Nonetheless, pseudo-labels still contain noise, which is progressively reduced during training.}
        \label{fig:cluster_init}
    \end{figure*}

    Since the pretrained model has never seen any of the private classes $C_P$ and the classifier $h_t$ is randomly initialised for its extended portion, most of the samples from private classes will be predicted into the shared classes. This will produce a certain amount of incorrect pseudo-labels, as shown in \Cref{fig:cluster_init}(a).
    To alleviate this problem, we propose a clustering-based solution to discover suitable prototypes for the private classes $C_P$, and thus initialize $W_P$ in a more convenient way. This will produce a better initial pseudo-labels assignment.
    The idea of using a clustering approach builds on the observation that the features of target samples (both from shared and private classes) are characterised by a low intra-class variability (\textit{c.f.} \Cref{fig:cluster_init}(a)). Therefore, a standard unsupervised clustering algorithm can , to some extent, group both shared and private target samples, providing the model with an improved initial pseudo-labeling. 
    Differently from other approaches \cite{open2, open3, aad}, this strategy avoids to rely on a separation criterion that hardly distinguishes whether samples belong to shared or target-private classes. 
    However, a major drawback in utilising an unsupervised clustering algorithm is the \textit{class misalignment}, as there is no correspondence between the clusters and the model classes and one needs to find the correct permutation which maps cluster ids to class ids.\footnote{In the classic example of \textit{cat vs. dog} classification, there is no guarantee that a clustering algorithm would assign cluster 1 to the class cat and cluster 2 to the class dog. As they can be easily inverted, such class misalignment becomes more exasperate in the presence of more than 2 classes.} 
    To alleviate this problem, we leverage the pretrained source model $g_s$ to align the shared classes $C_S$ with a subset of clusters obtained by the clustering algorithm. The remaining unmatched clusters are treated as unknown classes and their centroids as novel prototypes for the initialization of $W_P$, as depicted in \Cref{fig:cluster_init}. 

    More formally, we extract the features from the target samples $z_t = \varphi_s(x_t),\;\forall x_t \in \mathcal{X}_t$ using the pretrained source feature extractor $\varphi_s$. At this point, we perform an unsupervised clustering over all the extracted features $z_t$ to obtain $K$ centroids $c_k = \frac{1}{|\mathcal{S}_k|} \sum_{a \in \mathcal{S}_k} z_t^a$, where $k \in \{1,2,\ldots,K\}$, $K=|C_S|+|\widehat{C}_P|$, and $\mathcal{S}_k$ is the set of target samples assigned to the $k$-th cluster. Note that while $|\widehat{C}_S|$ is known, $|\widehat{C}_P|$ is arbitrarily chosen, since the real cardinality of the target private classes is unavailable under the open-set setting. In \cref{subsec:analysis}, we analyse the performance with different values for $|\widehat{C}_P|$ (which implies in different $K$).
    
    To account for the class misalignment, we align the shared classes with a subset of the obtained clusters by computing the cosine similarity between the shared class prototypes $W_S[i]$ and the cluster centroids $c_k$, 
    and we match each column in $W_S$ with the centroid with the highest similarity. This strategy solves the class misalignment by determining a permutation that reorders the centroids $c_k$ according to their match with shared class prototypes. 
    Finally, centroids that have not been assigned to any shared classes prototypes will be considered as discovered prototypes of the target-private classes, and used to fill the columns of $W_P$ (here the order of the clusters does not matter). 
    The refined model $W_T = [W_S|W_P]$ is then used to produce an initial pseudo-labels assignment, as detailed in the next section.

    
    \subsection{Pseudo-Label Refinement with Neighbours Consensus}
    \label{subsec:nnvoting}
    Similar to \cite{Litrico, contrastive_testtime}, the refinement of the pseudo-labels is accomplished by aggregating knowledge from nearest neighbour samples. The underlying idea is that similar samples are likely to belong to the same class. We thus assume that features extracted from semantically similar samples lie close to each other in the feature space and we further enforce this assumption by means of a contrastive learning loss, as described in \cref{subsec:temporal_queue_exclusion}. 
    
    More formally, given a target sample $x_t$ and a random weak augmentation $t_{wa}$ drawn from a distribution $\mathcal{T}_{wa}$, we obtain a feature vector $z_t = \varphi_t(t_{wa}(x_t))$ from the augmented image, which is used to search the neighbours of $x_t$ in the target features space \cite{Litrico}. The pseudo-label of $x_t$ is refined by aggregating predictions from the neighbours by soft-voting \cite{softvoting} as follows:
    \begin{equation}
        \label{eq:softvotingavg}
        \bar{p}_t^{(c)} = \frac{1}{|\mathcal{I}|}\sum_{i \in \mathcal{I}} p'^{(c)}_i,
    \end{equation}
    
    \noindent where $\mathcal{I}$ is the set of indices of the selected neighbours, $p'$ is the softmax output, and the superscript $c$ indicates the class index. 
    To obtain a refined pseudo-label, we find the  class $c$ with the highest probability computed in \cref{eq:softvotingavg}, \textit{i.e.} $\bar{y}_t = \argmax_c \; \bar{p}_t^{(c)}.$ 
    The refined pseudo-label is assigned to the sample $x_t$  and used as self-supervision signal (see \cref{subsec:overallframework}).
    
    This refining process relies on a bank $\mathcal{B}$ of length $M<|\mathcal{X}_t|$, storing pairs $\{z'_j, p_j'\}_{j=1}^M$ of features and softmax predictions. The neighbours of a sample $x_t$ are then selected by computing the cosine distances between the features stored in the bank.
    The $|\mathcal{I}|$ samples with the lowest distance are selected as neighbours.
    Following \cite{moco, Litrico}, we use a slowly changing momentum model $g'_t (\cdot) := \text{EMA}[h_t(\varphi_t(\cdot))]$ to update features $z'$ and predictions $p'$, in order to maintain the information stored in the bank stable during adaptation. A bank composed of $M$ randomly selected samples reduces the number of pairwise comparisons needed for finding the neighbours.

    \mypar{Bank initialisation with discovered classes prototypes}
    As described above, the refinement of the pseudo-labels is performed by averaging neighbours probabilities stored in $\mathcal{B}$. In order to generate the initial pseudo-labels that reflect the clusters obtained in \cref{subsec:initial_clustering}, we need to initialise the bank accordingly. However, the clustering strategy we use in \cref{subsec:initial_clustering} produces class assignments for all samples $x_t$ rather than probabilities. To overcome this issue, after the clustering initialisation, we associate to each target sample a vector of probabilities proportional to its similarity to the centroids $c_k$.
    Given target features $z_t$ and centroids $c_k$, we determined the probabilities vectors $p'$ as the per-class similarity between $z_t$ and each centroid in $c_k$. Formally, for each target sample, we obtain the probabilities vector $p'$ to initialise the bank as follows: 
    \begin{equation}
        p' = 1 - \frac{CosDist(z_t, c_k)}{\underset{j \in \{1,...,|\hat{C}_P|\}}{\max} CosDist(z_t, c_k^j)} 
    \end{equation}
    
    \noindent where $CosDist(z_t, c_k) = 1 - \frac{z_t \cdot c_k}{||z_t||\;||c_k||}$ is the cosine distance and $i$ is the class index. Finally, to obtain probabilities, the vector $p'$ is provided to the softmax function $\sigma$ with a temperature parameter $\tau_2$, $p' = \sigma (p' / \tau_2)$.
    Consequently, for all samples $x_t$ the probability value for class $c$ will be proportional to the similarity between the features $z_t = \varphi_t(x_t)$ and the centroid of $c$. Also, the class with the highest probability will be exactly the one assigned to the sample by the clustering strategy in \cref{subsec:initial_clustering}.

    \subsection{Pseudo-labels Uncertainty in the Open-set Scenarios}
    \label{subsec:entropy_estimation}

    The refined pseudo-labels $\bar{y}_t$ are used as a self-supervision signal for a classification loss on target data, as detailed in \cref{subsec:overallframework}.
    However, such pseudo-labels are far from perfect, especially at the beginning of the iterative refining process, and thus still contain noise. Therefore, using all of them in the classification loss will disrupt adaptation, since the model would be trained with incorrect information. To mitigate this, we perform a sample selection, in order to pick only samples with low uncertainty: these are selected stochastically via Bernoulli sampling, based on the uncertainty of their pseudo-label, estimated in a twofold manner, as detailed below. 

    \mypar{Uncertainty estimation via Neighbours Consensus} The first uncertainty estimation method is as our previous work \cite{Litrico}. It estimates the uncertainty of pseudo-labels (after their refinement) by measuring the consensus among neighbours' predictions. This strategy builds upon the intuition that if the neighbourhood agrees on a predicted class, the derived pseudo-labels should be considered reliable (low uncertainty). Otherwise, if the network predicts different classes for the neighbour samples, the obtained pseudo-labels should be considered unreliable (high uncertainty).
    To this end, we compute the entropy $\mathcal{H}$ of $\bar{p}_t$ (\cref{eq:softvotingavg}) as an estimator of pseudo-label uncertainty, noting that $\mathcal{H}$ yields low values in low uncertainty cases, and high values otherwise.
    We define the uncertainty coefficient $u^{nc}_t$  as follows:
    \begin{equation}
        u^{nc}_t = \frac{\mathcal{H}(\bar{p}_t)}{\log_2 \, |C|}.
    \end{equation}
    While the uncertainty coefficient $u^{nc}_t$ reduces the impact of the noisy pseudo-labels in a closed-set scenario \cite{Litrico}, we noted that it can fail to model uncertainty in open-set scenario. In fact, during the initial clustering assignment, a cluster made of target-private samples can be incorrectly assigned to any of the shared classes. This is mainly due to the fact that the model has not been optimised to separate shared and target-private classes during the pretraining on the source, implying that shared and target-private clusters can lie very close in the feature space.
    In this scenario, all the samples of the clusters will share the same incorrect pseudo-label, making the uncertainty estimation unsuccessful.  

\begin{figure}[t]
    \centering
        \includegraphics[width=\linewidth]{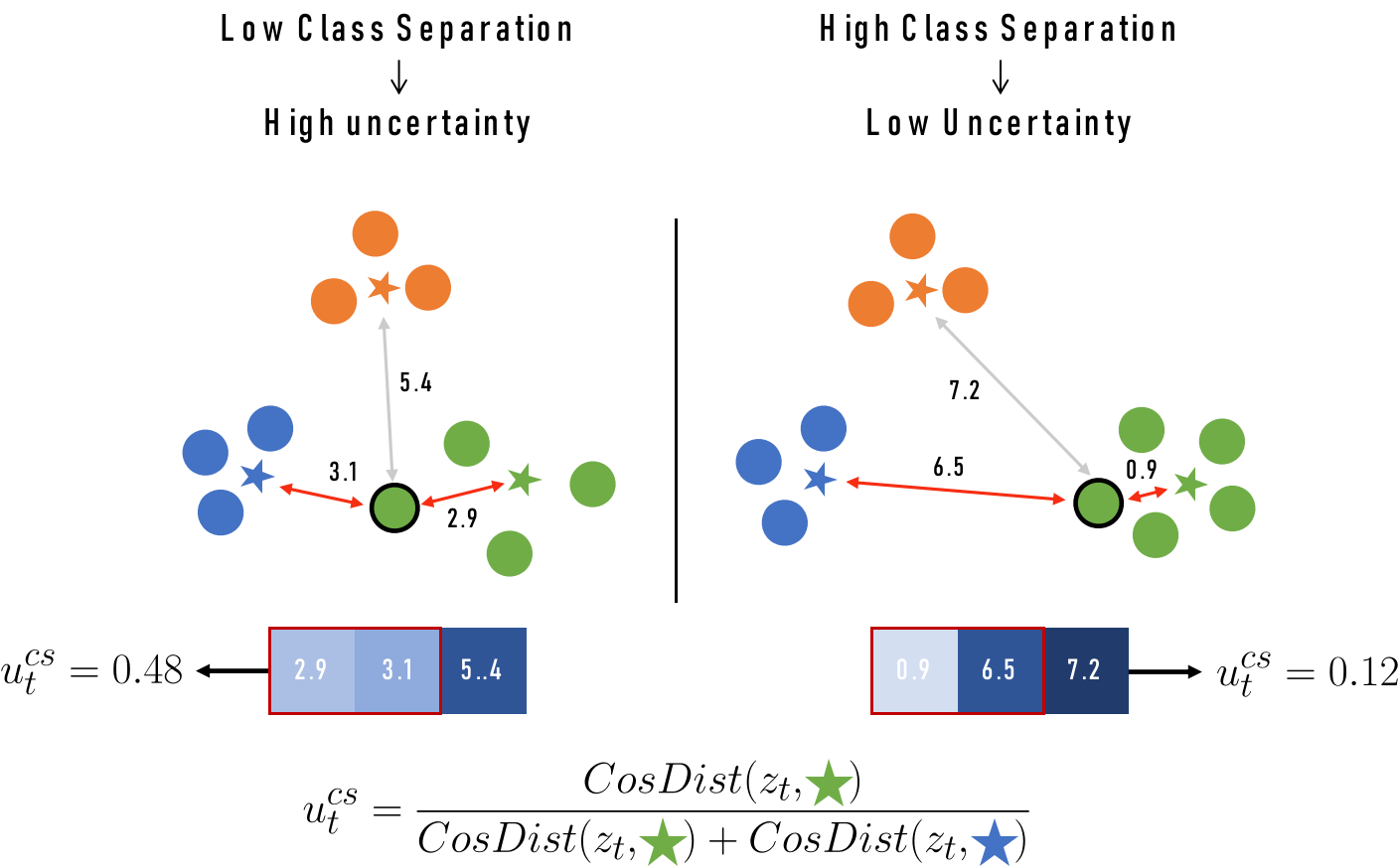}
        \caption{\label{fig:openset-weighting} Pseudo-label uncertainty via Class Separation. The uncertainty of the pseudo-labels is estimated by computing the distance of samples from the two closest class prototypes. If a sample is similarly distant in the features space from two classes, its pseudo-label has high uncertainty. Contrarily, if a sample is close to one class and far from the other, its pseudo-label has low uncertainty and we consider it as reliable.}
    \end{figure}

    \mypar{Uncertainty estimation via Class Separation}
    To tackle this problem, we propose a novel strategy to estimate the uncertainty of pseudo-labels that is well-suited in a open-set scenario, by analysing the distance of samples to class prototypes, as depicted in \cref{fig:openset-weighting}.
    We build upon the intuition that when two classes are close in the features space, the distance of the samples from the two classes prototypes will be similar. Contrarily, when those classes are clearly separated, the samples will be close to one prototype and far from the other. Following this intuition, we consider the former and the latter scenarios of high and low uncertainty, respectively. 
    For instance, in a situation where a cluster of target-private samples is so close to a shared class that can be switched, the distances of samples to prototypes of the two classes will be similar, resulting in high uncertainty.
    More formally, for a target sample $x_t$ we select from $\bar{p}_t$ the indexes $\{i,j\}$ of the two classes with the largest values as the most probable pseudo-labels for the sample. Then we compute the cosine distance between the features of the target sample $z_t = \varphi_t(x_t)$ and the prototypes $W[i]$ and $W[j]$. 
    Finally, we obtain the uncertainty value $u^{cs}_t$ as follows:
    \begin{equation}
        u^{cs}_t = \frac{CosDist(z_t, W[i])}{CosDist(z_t, W[i]) + CosDist(z_t, W[j])}
    \end{equation}

    \mypar{Bernoulli Sampling} 
    To improve the model robustness to the noisy pseudo-labels for samples of both shared and target-private classes, we combine these two uncertainty estimation strategies for selecting samples with the lowest uncertainty to be used in the classification loss, as detailed in \cref{subsec:overallframework}.
    Specifically, we compute the probabilities $w^{nc}_t$ and $w^{cs}_t$ for a target sample to have a correct pseudo-label with respect to both $u^{nc}_t$ and $u^{cs}_t$, respectively:

    \begin{equation}
        \label{eq:weights}
        w^{nc}_t = \mathcal{F}(u^{nc}_t)
    \end{equation}
    \begin{equation}
        \label{eq:weights2}
        w^{cs}_t = \mathcal{F}(u^{cs}_t)
    \end{equation}
    where $\mathcal{F}$ is a monotonically decreasing function to assign high probabilities to samples with low uncertainty and viceversa.  In \cref{subsec:analysis}, we analyse two choices for $\mathcal{F}$.

    
    We select low uncertainty samples via Bernoulli sampling with probabilities $w^{nc}_t$ and $w^{cs}_t$, which are combined as follows:
    \begin{equation}
    \label{eq:merging_uncertainty}
    M_i = \operatorname{B}(w^{nc}_t) \oplus \operatorname{B}(w^{cs}_t) 
    \end{equation}    

    \noindent where $\operatorname{B}(p) \in \left\lbrace0,1\right\rbrace$ is the Bernoulli sampling with probability of success $0\leq p\leq 1$ and $\oplus$ is a logical operator. In \cref{subsec:analysis} we compare two choices for the operator $\oplus$.
    
    Finally, we define $\mathcal{U}=\{x_t^i \in \mathcal{X}_t| M_i = 1\}$ as the set of uncertainty-based selected target samples. This set of samples is used during the adaptation process in the classification loss, as detailed in the next section.

    \begin{figure}[t!]
    \includegraphics[width=\columnwidth]{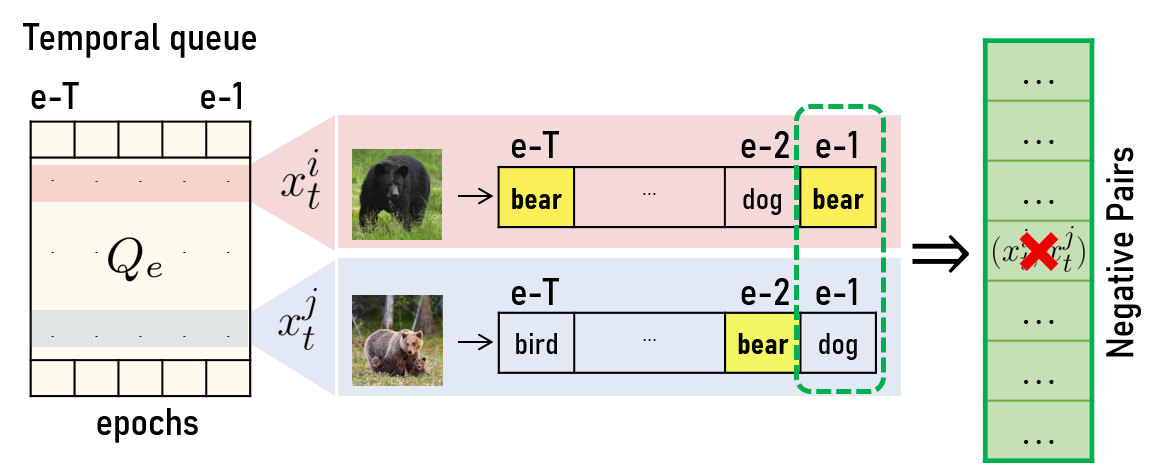}
        \vspace{-10pt}
    \centering
        \vspace{-8pt}
    \caption{A couple of target images would be wrongly considered as a negative pair if only comparing the latest predictions (green box). Instead, since $x_t^i$ and $x_t^j$ share the same pseudo-labels (at least once) in the past T epochs, i.e. $\{e-T,...,e-1 \}$, we exclude them from the list of negative pairs. Figure from \cite{Litrico}, reproduced with the permission of the authors.} 
    \label{fig:queue_based_exclusion}
    \vspace{-10pt}
    \end{figure}
    
    \subsection{Adaptation with Refined Pseudo-labels}
    \label{subsec:adapt-pseudlabel}
    The refined pseudo-labels obtained with neighbours' knowledge aggregation are used to compute a classification loss on target data to adapt the model to the new domain.
    We use the refined pseudo-labels $\bar{y}_t$ obtained from a weakly-augmented image $t_\text{wa}(x_t)$ as self-supervision for a strongly-augmented version $t_\text{sa}(x_t)$. The refining of the pseudo-labels is an iterative process, progressively improving the pseudo-labels accuracy during training through several refining steps. 
    
    As in our previous work \cite{Litrico}, in addition to the proposed sample selection and exclusion strategies, to mitigate the effect of the noisy pseudo-labels, we use the negative learning loss \cite{NLNL, JPNL} as classification loss.  Differently from \cite{NLNL, JPNL}, which use the negative learning loss concurrently or alternating with a positive loss, here we do not use a positive loss in the entire training. 
    The used classification loss is the following:
    \begin{equation}
    \label{eq:cls_loss}
    L_t^{cls} = - \; \mathbb{E}_{x_t \in \mathcal{U}}  \Bigr[ \sum_{c=1}^C  \tilde{y}^c \, log \, (1-p_\text{sa}^c) \Bigr],
    \end{equation}
    where $\tilde{y}^c$ is a complementary label $\tilde{y} \in \{1,\ldots,C\} \: \backslash \: \{\bar{y}_t\}$ chosen randomly from the set of labels and without the refined pseudo-labels, $p_\text{sa} = \sigma(h_t(\varphi_t(t_\text{sa}(x_t))))$ is the probabilistic output for the strongly-augmented image $t_\text{sa}(x_t)$, and $\mathcal{U}$ is the set of samples with low uncertainty, as detailed in \Cref{subsec:entropy_estimation}.
    The random selection of the complementary label $\tilde{y}$ is coherent with the negative learning framework \cite{NLNL}.
    

        \begin{table*}[t!]
\caption{HOS score (\%) on Office31. All methods use the ResNet-50 backbone. The proposed approach outperforms the previous SF-UDA state-of-the-art by 2.1\% on average (Avg.), while being competitive with UDA baselines. The underlined results are the best within UDA methods, while the bold one are the best SF-UDA approaches.}
\begin{adjustbox}{width=\textwidth}
\begin{tabular}{lc||ccc|ccc|ccc|ccc|ccc|ccc||ccc}
\toprule

\multicolumn{23}{c}{Office31} \\ \midrule
\multirow{2}{*}{Method} & \multirow{2}{*}{Source-free} & \multicolumn{3}{c|}{A $\rightarrow$ D}  & \multicolumn{3}{c|}{A $\rightarrow$ W} & \multicolumn{3}{c|}{D $\rightarrow$ A}  & \multicolumn{3}{c|}{D $\rightarrow$ W}  & \multicolumn{3}{c|}{W $\rightarrow$ A} & \multicolumn{3}{c||}{W $\rightarrow$ D} & \multicolumn{3}{c}{\textbf{Average}} \\
 & & OS* & UNK & \textbf{HOS} & OS* & UNK & \textbf{HOS} & OS* & UNK & \textbf{HOS} & OS* & UNK & \textbf{HOS} & OS* & UNK & \textbf{HOS} & OS* & UNK & \textbf{HOS} & OS* & UNK & \textbf{HOS} \\ \midrule
STA \cite{open3} & \xmark & 91.0 & 63.9 & 75.0 & 86.7 & 67.6 & 75.9 & 83.1 & 65.9 & 73.2 & 94.1 & 55.5 & 69.8 & 66.2 & 68.0 & 66.1 & 84.9 & 67.8 & 75.2 & 84.3 & 64.8 & 72.6 \\
OSBP \cite{open4} & \xmark & 90.5 & 75.5 & 82.4 & 86.8 & 79.2 & 82.7 & 76.1 & 72.3 & 75.1 & 97.7 & 96.7 & 97.2 & 73.0 & 74.4 & 73.7 & 99.1 & 84.2 & 91.1 & 87.2 & 80.4 & 83.7 \\
UAN \cite{uan} & \xmark & \underline{95.6} & 24.4 & 38.9 & \underline{95.5} & 31.0 & 46.8 & 93.5 & 53.4 & 68.0 & \underline{99.8} &  52.5 & 68.8 & \underline{94.1} & 38.8 & 54.9 & 81.5 & 41.4 & 53.0 & \underline{93.4} & 40.3 & 55.1 \\
OSLPP \cite{open5} & \xmark & 92.6 & 90.4 & \underline{91.5} & 89.5 & 88.4 & \underline{89.0} & 82.1 & 76.6 & 79.3 & 96.9 & 88.0 & 92.3 & 78.9 & 78.5 & 78.7 & 95.8 & 91.5 & 93.6 & 89.3 & 85.6 & 87.4 \\
ROS \cite{open1} & \xmark & 87.5 & 77.8 & 82.4 & 88.4 & 76.7 & 82.1 & 74.8 & 81.2 & 77.9 & 99.3 & 93.0 & 96.0 & 69.7 & 86.6 & 77.2 & \underline{100.0} & \underline{99.4} & \underline{99.7} & 86.6 & 85.8 & 85.9 \\
DANCE \cite{self_supervised7} & \xmark & 94.3 & 50.7 & 66.9 & 92.8 & 55.9 & 70.7 & 97.0 & 66.8 & 80.0 & 97.6 & 73.7 & 84.8 & 82.4 & 85.3 & 65.8 & 81.6 & 60.6 & 70.2 & 91.0 & 60.2 & 73.1 \\
cUADAL \cite{open2} & \xmark & 86.4 & \underline{95.1} & 90.1 & 86.0 & \underline{90.4} & 87.9 & \underline{98.6} & \underline{97.7} & \underline{98.2} & 99.3 & 99.4 & 99.4 & 75.4 & 87.8 & 80.5 & 67.6 & 87.8 & 75.1 & 85.6 & \underline{93.0} & 88.5 \\
ANNA \cite{anna} & \xmark & 93.2 & 76.1 & 83.8 & 82.8 & 88.4 & 85.5 & 75.4 & 91.1 & 82.5 & 99.4 & \underline{99.6} & \underline{99.5} & 76.0 & \underline{87.9} & \underline{81.6} & \underline{100.0} & 96.8 & 98.4 & 87.8 & 90.0 & \underline{88.6} \\ \midrule 

SHOT \cite{sfda2} & \cmark & \textbf{94.0} & 46.3 & 62.0 & \textbf{95.6} & 42.3 & 58.7 & \textbf{83.3} & 39.1 & 53.3 & \textbf{100.0} & 75.7 & 86.1 & \textbf{82.7} & 46.6 & 59.6 & \textbf{100.0} & 69.7 & 82.1 & \textbf{92.6} & 53.3 & 67.0 \\
AaD \cite{aad} & \cmark & 73.0 & 84.6 & 78.3 & 63.5 & \textbf{89.5} & 74.3 & 63.6 & \textbf{88.9} & 74.2 & 78.0 & \textbf{98.5} & 87.0 & 61.9 & \textbf{88.9} & 73.0 & 94.6 & \textbf{96.8} & 95.7 & 72.4 & \textbf{91.2} & 80.4 \\
USD \cite{usd} & \cmark & 90.7 & 73.4 & 81.2 & 82.8 & 72.7 & 77.9 & 65.7 & 84.4 & 73.9 & 97.9 & 96.6 & \textbf{97.3} & 64.6 & 86.7 & \textbf{74.0} & 98.0 & 92.6 & 95.2 & 83.3 & 84.4 & 83.3 \\
\textbf{Ours} & \cmark & 85.7 & \textbf{93.0} & \textbf{89.2}  & 90.9 & 77.8 & \textbf{83.9}  & 73.3 & 83.4 & \textbf{78.0}  & 89.9 & 95.7 & 93.2  & 65.0 & 80.0 & 71.8 & 98.6 & 94.6 & \textbf{96.6} & 83.9 & 89.6 & \textbf{85.4} \\ \bottomrule
                                          
\end{tabular}
\end{adjustbox}
\label{tab:office31}
\end{table*}
    
    \subsection{Contrastive Loss for Open-set Scenarios}
    \label{subsec:temporal_queue_exclusion}
    As described in \cref{subsec:nnvoting}, the process of pseudo-labels refinement through neighbours samples assumes that features extracted from same-class samples lie closer in the features space, compared to samples from other classes. We encourage this assumption during adaptation via self-supervised contrastive learning. We build on MoCo~\cite{moco} to aggregate features from different augmentations of the same image and separate representations originated from different samples (negative pairs). Specifically, for each sample $x_t$, we generate two strongly-augmented samples $t_{sa}(x_t)$ and $t'_{sa}(x_t)$, where $t_{sa}, t'_{sa} \in \mathcal{T}_\text{sa}$ are two randomly sampled transformations from the space of strong augmentations $\mathcal{T}_{sa}$. To build the constrastive pairs in the feature space, the feature extractor embeds augmented samples as query $q = \varphi_t(t_{sa}(x_t))$ and key $k = \varphi_t(t_{sa}(x_t))$. While we employ representations $q$ and $k$ associated to the same sample $x_t$ as positive pairs, we build negative pairs by maintaining a queue $Q_e$ that stores key features $\{ k^{(i)}\}_{i=1}^N$ as detailed below.
        
    MoCo \cite{moco} uses the pairs $(q,k)$ as positive and all the pairs $\{(q,k^{(i)}) \}_{i=1}^N$ as negative pairs by minimising and maximising their cosine distance for positive and negative pairs, respectively. Since features are stored in $Q_e$ independently from the class, even features from samples belonging to the same class will be pushed away, which is in contrast with our objective to better aggregate same-class samples in the feature space.
    In \cite{contrastive_testtime}, authors propose a strategy to exclude some negative pairs from the contrastive loss. For every negative pair, they just compare the pseudo-labels of the two samples. If they share the same pseudo-label, then the negative pair is excluded; otherwise, the pair is included in the negative pairs list. 
    However, the exclusion strategy in \cite{contrastive_testtime} does not take into account the noise that inevitably affects the pseudo-labels during adaptation. 
    Indeed, such noise may still lead to wrong negative pairs, \textit{i.e.}, pairs classified differently but having the same ground-truth label. This causes instability of standard contrastive frameworks, highly affecting the learned features space. Hence, we follow the temporal exclusion strategy in ~\cite{Litrico}, leveraging past predictions to identify and exclude pairs composed of same class samples, even with noisy pseudo-labels.
    This exclusion strategy is based on the intuition that, rather than only looking at the current pseudo-label, it is more reliable to look at its past history to have a higher probability of observing, at least once, the correct label: the history will probably reveal the correct one, improving the exclusion process.
    To this end, we build a temporal queue $Q_e$ by storing, for each sample, also the refined pseudo-labels $\{\bar{y}^{(j)}\}_{j=1}^\tau$ of the $\tau$ past epochs, \textit{i.e.}, $\{e-\tau,\ldots,e-1\}$, and we filter out all the pairs that shared the same pseudo-labels at least once in the past $\tau$ epochs from the set of available negative pairs, as illustrated in \cref{fig:queue_based_exclusion}. Although this process will likely filter out a lot of true negatives, it ensures most of the false negatives are excluded.

    \mypar{NL-InfoNCELoss}
    The refining of the pseudo-labels is an iterative process, which implies that an amount of noise is still present in the pseudo-labels. Moreover, the presence of target-private classes in the open-set setting, which are hard to be correctly identified, increases such noise with respect to the closed-set case. Inspired by \cite{NLNL} and differently from our previous work \cite{Litrico}, we propose a novel contrastive loss, called NL-InfoNCELoss, that inserts the principle of negative learning in the standard InfoNCELoss. Although the strategies proposed in \cref{subsec:entropy_estimation} and \cref{subsec:temporal_queue_exclusion} aim at reducing the impact of such noise, we posit that leveraging the negative learning paradigm can be beneficial for a contrastive loss to reduce the impact of false negative samples.  
    
    Accordingly, the formulation of the proposed NL-InfoNCELoss is the following:
        \begin{equation}
    L_{\text{NL-InfoNCE}} = - \, log \left( 1 - \, \frac{exp(q \cdot {k}_{-} / \tau)}{\sum_{j\in \mathcal{N}_q} exp(q \cdot k_j / \tau)} \right),
        \end{equation}
    $$
    \mathcal{N}_q = \{ j | \bar{y}_j^{(i)} \neq \bar{y}^{(i)} , \;\forall j \in \{1,\ldots,N\}, \forall i\in\{1,\ldots,\tau\} \},
    $$
    \noindent where ${k}_{-}$ is one negative sample chosen randomly from the set of negative samples, $\mathcal{N}_q$ is the set of indices of samples in $Q_e$ that never shared with the query sample the same pseudo-labels in the past $\tau$ epochs. By optimising the proposed NL-InfoNCELoss, the model is trained to increase the feature distance between the query and a randomly selected negative sample, rather than to all the negative samples, as in the standard InfoNCELoss. Albeit this formulation slackens the separation effect of the contrastive loss, it reduces the impact of noisy pseudo-labels in the negative pairs.
    
    In \cref{subsec:analysis}, we analyse the contribution of the proposed loss compared to the standard InfoNCELoss.

    \subsection{Overall Framework}
    \label{subsec:overallframework}
    In addition to the classification and the contrastive losses detailed in the previous sections, following the standard state-of-the-art protocol \cite{Litrico},  we add the following regularisation term to prevent the posterior collapse: 
    \begin{equation*}
        L_t^{div} = \mathbb{E}_{x_t \in \mathcal{X}_t} \Bigr[ \sum_{c=1}^C  \bar{p}_q^c \, log \, \bar{p}_q^c \Bigl],
    \end{equation*}
    \begin{equation*}
        \bar{p}_q = \mathbb{E}_{x_t \in \mathcal{X}_t}  \left[ \sigma\left(h_t\left(\varphi_t\left(t_{sa}\left(x_t\right)\right)\right)\right)  \right].
        \end{equation*}

    \noindent When trained with noisy labels, the model may learn a degenerate latent representation that causes to predict all the samples in a single class, especially if the noise is very skewed towards a single category. By encouraging diversity in model predictions, this term helps to avoid such phenomenon.
        
    The overall objective function used to adapt the model to the target data is the following:
        \begin{equation}
        L_t = \gamma_1 L_t^{cls} + \gamma_2 L_t^{ctr} + \gamma_3 L_t^{div},
        \end{equation}

\begin{table*}[t]
\caption{HOS score (\%) on Office-Home. All methods use the ResNet-50 backbone. The proposed approach achieves the highest accuracy on average (Avg.) compared with SF-UDA baselines and outperforms several UDA methods. The underlined results are the best within standard UDA methods, while bold ones are the best SF-UDA approaches.}
\begin{adjustbox}{width=\textwidth}
\begin{tabular}{lc||ccc|ccc|ccc|ccc|ccc|ccc}
\toprule

\multicolumn{20}{c}{Office-Home} \\ \midrule
\multirow{2}{*}{Method} & \multirow{2}{*}{Source-free} & \multicolumn{3}{c|}{Ar $\rightarrow$ Cl}  & \multicolumn{3}{c|}{Ar $\rightarrow$ Pr} & \multicolumn{3}{c|}{Ar $\rightarrow$ Rw}  & \multicolumn{3}{c|}{Cl $\rightarrow$ Ar}  & \multicolumn{3}{c|}{Cl $\rightarrow$ Pr} & \multicolumn{3}{c}{Cl $\rightarrow$ Rw} \\
 & & OS* & UNK & \textbf{HOS} & OS* & UNK & \textbf{HOS} & OS* & UNK & \textbf{HOS} & OS* & UNK & \textbf{HOS} & OS* & UNK & \textbf{HOS} & OS* & UNK & \textbf{HOS}\\ \midrule
PGL \cite{pgl} & \xmark & 63.3 & 19.1 & 29.3 & 78.9 & 32.1 & 45.6 & \underline{87.7} & 40.9 & 55.8 & \underline{85.9} & 5.3 & 10.0 & 73.9 & 24.5 & 36.8 & 70.2 & 33.8 & 45.6 \\
STA \cite{open3} & \xmark & 50.8 & 63.4 & 56.3 & 68.7 & 59.7 & 63.7 & 81.1 & 50.5 & 62.1 & 53.0 & 63.9 & 57.9 & 61.4 & 63.5 & 62.5 & 69.8 & 63.2 & 66.3 \\
OSBP \cite{open4} & \xmark & 50.2 & 61.1 & 55.1 & 71.8 & 59.8 & 65.2 & 79.3 & 67.5 & 72.9 & 59.4 & 70.3 & 64.3 & 67.0 & 62.7 & 64.7 & 72.0 & 69.2 & 70.6 \\
DAOD \cite{daod} & \xmark & \underline{72.6} & 51.8 & 60.5 & 55.3 & 57.9 & 56.6 & 78.2 & 62.6 & 69.5 & 59.1 & 61.7 & 60.4 & 70.8 & 52.6 & 60.4 & 77.8 & 57.0 & 65.8 \\
OSLPP \cite{open5} & \xmark & 55.9 & 67.1 & 61.0 & 72.5 & 73.1 & 72.8 & 80.1 & 69.4 & 74.3 & 49.6 & 79.0 & 60.9 & 61.6 & 73.3 & 66.9 & 67.2 & 73.9 & 70.4 \\
ROS \cite{open1} & \xmark & 50.6 & 74.1 & 60.1 & 68.4 & 70.3 & 69.3 & 75.8 & 77.2 & 76.5 & 53.6 & 65.5 & 58.9 & 59.8 & 71.6 & 65.2 & 65.3 & 72.2 & 68.6 \\
DANCE \cite{self_supervised7} & \xmark & 54.4 & 53.7 & 53.1 & \underline{82.2} & 35.4 & 49.8 & 87.4 & 25.3 & 39.4 & 71.2 & 28.4 & 40.9 & \underline{74.6} & 32.8 & 45.9 & \underline{81.3} & 18.4 & 30.2 \\
cUADAL \cite{open2} & \xmark & 55.8 & 75.6 & 63.6 & 69.6 & 73.9 & 71.6 & 81.8 & 73.3 & \underline{77.5} & 54.9 & \underline{82.0} & \underline{65.0} & 61.8 & \underline{77.4} & 68.3 & 69.5 & 76.3 & 72.6 \\
ANNA \cite{anna} & \xmark & 61.4 & \underline{78.7} & \underline{69.0} & 68.3 & \underline{79.9} & \underline{73.7} & 74.1 & \underline{79.7} & 76.8 & 58.0 & 73.1 & 64.7 & 64.2 & 73.6 & \underline{68.6} & 66.9 & \underline{80.2} & \underline{73.0} \\ \midrule

SHOT \cite{sfda2} & \cmark & \textbf{67.0} & 28.0 & 39.5 & \textbf{81.8} & 26.3 & 39.8 & \textbf{87.5} & 32.1 & 47.0 & \textbf{66.8} & 46.2 & 54.6 & \textbf{77.5} & 27.2 & 40.2 & \textbf{80.0} & 25.9 & 39.1 \\
AaD \cite{aad} & \cmark & 50.7 & 66.4 & 57.6 & 64.6 & 69.4 & 66.9 & 73.1 & 66.9 & 69.9 & 48.2 & \textbf{81.1} & \textbf{60.5} & 59.5 & 63.5 & 61.4 & 67.4 & 68.3 & 67.8 \\
USD \cite{usd} & \cmark & 53.3 & 71.5 & \textbf{61.1} & 65.7 & 74.9 & \textbf{70.0} & 73.3 & 79.5 & 76.3 & 52.2 & 70.8 & 60.1 & 62.4 & 68.4 & 65.2 & 69.3 & 68.6 & 68.9 \\
\textbf{Ours} & \cmark & 46.6 & \textbf{72.8} & 56.9 & 61.1 & \textbf{80.0} & 69.3 & 74.1 & \textbf{84.6} & \textbf{79.0} & 44.8 & 62.4 & 52.7 & 62.7 & \textbf{69.6} & \textbf{66.0} & 69.5 & \textbf{76.9} & \textbf{73.0} \\ \bottomrule
                                          
\end{tabular}
\end{adjustbox}
\vspace{-5pt}
\label{tab:officehome}
\end{table*}

\begin{table*}[ht]
\begin{adjustbox}{width=\textwidth}
\begin{tabular}{l||ccc|ccc|ccc|ccc|ccc|ccc||ccc}
\multirow{2}{*}{Method} & \multicolumn{3}{c|}{Pr $\rightarrow$ Ar}  & \multicolumn{3}{c|}{Pr $\rightarrow$ Cl} & \multicolumn{3}{c|}{Pr $\rightarrow$ Rw}  & \multicolumn{3}{c|}{Rw $\rightarrow$ Ar}  & \multicolumn{3}{c|}{Rw $\rightarrow$ Cl} & \multicolumn{3}{c||}{Rw $\rightarrow$ Pr} & \multicolumn{3}{c}{\textbf{Average}} \\
& OS* & UNK & \textbf{HOS} & OS* & UNK & \textbf{HOS} & OS* & UNK & \textbf{HOS} & OS* & UNK & \textbf{HOS} & OS* & UNK & \textbf{HOS} & OS* & UNK & \textbf{HOS} & OS* & UNK & \textbf{HOS}\\ \midrule

PGL \cite{pgl} & \underline{73.7} & 34.7 & 47.2 & \underline{59.2} & 38.4 & 46.6 & \underline{84.8} & 27.6 & 41.6 & \underline{81.5} & 6.1 & 11.4 & \underline{68.8} & 0.0 & 0.0 & \underline{84.8} & 38.0 & 52.5 & \underline{76.1} & 25.0 & 35.2 \\
STA \cite{open3} & 54.2 & 72.4 & 61.9 & 44.2 & 67.1 & 53.2 & 76.2 & 64.3 & 69.5 & 67.5 & 66.7 & 67.1 & 49.9 & 61.1 & 54.5 & 77.1 & 55.4 & 64.5 & 61.8 & 63.3 & 61.1 \\
OSBP \cite{open4} & 59.1 & 68.1 & 63.2 & 44.5 & 66.3 & 53.2 & 76.2 & 71.7 & 73.9 & 66.1 & 67.3 & 66.7 & 48.0 & 63.0 & 54.5 & 76.3 & 68.6 & 72.3 & 64.1 & 66.3 & 64.7 \\
DAOD \cite{daod} & 71.3 & 50.5 & 59.1 & 58.4 & 42.8 & 49.4 & 81.8 & 50.6 & 62.5 & 66.7 & 43.3 & 52.5 & 60.0 & 36.6 & 45.5 & 84.1 & 34.7 & 49.1 & 69.6 & 50.2 & 57.6 \\
OSLPP \cite{open5} &  54.6 & 76.2 & 63.6 & 53.1 & 67.1 & 59.3 & 77.0 & 71.2 & 74.0 & 60.8 & 75.0 & 67.2 & 54.4 & 64.3 & 59.0 & 78.4 & 70.8 & 74.4 & 63.8 & 71.7 & 67.0 \\
ROS \cite{open1} & 57.3 & 64.3 & 60.6 & 46.5 & 71.2 & 56.3 & 70.8 & 78.4 & 74.4 & 67.0 & 70.8 & 68.8 & 51.5 & 73.0 & 60.4 & 72.0 & 80.0 & 75.7 & 61.6 & 72.4 & 66.2 \\
DANCE \cite{self_supervised7} & 69.7 & 43.9 & 54.2 & 48.9 & 67.4 & 55.7 & 84.2 &  27.1 & 41.2 & 76.8 & 16.7 & 27.5 & 59.4 & 41.3 & 48.3 & 84.1 & 29.6 & 44.0 & 72.8 & 35.0 & 44.2\\
cUADAL \cite{open2} & 52.1 & \underline{82.4} & 62.9 & 42.7 & \underline{80.7} & 54.6 & 71.7 & \underline{83.4} & \underline{76.8} & 67.3 & \underline{79.6} & \underline{72.6} & 52.5 & 71.1 & 59.9 & 77.7 & 75.6 & 76.7 & 63.1 & \underline{77.6} & 68.5\\
ANNA \cite{anna} & 63.0 & 70.3 & \underline{66.5} & 54.6 & 74.8 & \underline{63.1} & 74.3 & 78.9 & 76.6 & 66.1 & 77.3 & 71.3 & 59.7 & \underline{73.1} & \underline{65.7} & 76.4 & \underline{81.0} & \underline{78.7} & 65.6 & 76.7 & \underline{70.7} \\ \midrule

SHOT \cite{sfda2} & \textbf{66.3} & 51.1 & 57.7 & 59.3 & 31.0 & 40.8 & \textbf{85.8} & 31.6 & 46.2 & \textbf{73.5} & 50.6 & 59.9 & \textbf{65.3} & 28.9 & 40.1 & \textbf{84.4 }& 28.2 & 42.3 & \textbf{74.6} & 33.9 & 45.6 \\
AaD \cite{aad} & 47.3 & \textbf{82.4} & 60.1 & 45.4 & 72.8 & 55.9 & 68.4 & 72.8 & 70.6 & 54.5 & \textbf{79.0} & 64.6 & 49.0 & 69.6 & 57.5 & 69.7 & 70.6 & 70.1 & 58.2 & 71.9 & 63.6 \\
USD \cite{usd} & 54.3 & 73.8 & \textbf{62.6} & 47.3 & 69.6 & 56.3 & 70.0 & 74.5 & 72.2 & 64.6 & 71.3 & \textbf{67.8} & 53.8 & 65.5 & 59.1 & 73.3 & 69.1 & 71.1 & 61.6 & 71.5 & 65.9 \\
\textbf{Ours} & 52.3 & 63.7 & 57.4 & \textbf{65.3} & \textbf{80.7} & \textbf{72.2} & 66.8 & \textbf{78.6} & \textbf{72.3} & 51.4 & 77.4 & 61.8 & 50.3 & \textbf{74.6} & \textbf{60.1} & 69.7 & \textbf{73.1} & \textbf{71.4} & 59.5 & \textbf{74.5} & \textbf{66.0}\\ \bottomrule
                                          
\end{tabular}

\end{adjustbox}
\label{tab:officehome2}
\end{table*}

\section{Experiments and Results}
\label{sec:experiments_results}
    \mypar{Datasets} To evaluate our approach, we performed experiments on Office31 \cite{office31} and Office-Home \cite{officehome} for the open-set source-free domain adaptation setting. \\
    \textit{\textbf{\textendash~ Office31} } contains three domains, Amazon (A), Dslr (D), and Webcam (W), with 31 classes, where the first 10 classes are shared and the last 11 private classes. The remaining 10 classes are not used in this setting.\\
    \textit{\textbf{\textendash~ Office-Home} } consists of 65 classes of labeled images deriving from four specific domains, Art (Ar), Clipart (Cl), Product (Pr), and Real World (Rw). The first 25 categories in alphabetic order are shared classes and the remaining 40 classes are private.\\
    \mypar{Evaluation metrics} Following the literature \cite{anna, open1, open3}, we use three evaluation metrics to compare the performance, including the average class accuracy over known classes $\text{OS}^{\ast}$, the accuracy of unknown class UNK, and $\text{HOS} = 2 \, \frac{\text{OS}^{\ast} \times \text{UKN}}{\text{OS}^{\ast} + \text{UNK}}$ which is the harmonic mean between OS$^{\ast}$ and UNK. HOS is the core metric in the latest open-set domain adaptation literature, since it requires a good balance on both shared and private class accuracy in an unbiased manner.\\
    \mypar{Implementation details} We use standard classification architectures comprising a feature extractor followed by a classifier. For fair comparison purposes, we choose the ResNet-50 model \cite{resnet} used by competitors as a backbone in the experiments.  
 
    For source training, we initialise the ResNet backbone with ImageNet-1K \cite{imagenet} pre-trained weights available in the Pytorch model zoo. We train the source model with \textit{only a standard cross-entropy loss}. For the adaptation stage, the target model is initialised with the source model's parameters. For open-set experiments, before the adaptation, we increase the classifier size to $2 \cdot |C_S|$ to account for the unknown classes. 
    The initial clustering is performed with the K-means algorithm of Scikit-learn \cite{scikit-learn}.
    For more details, the code will be made available upon paper acceptance.
    

    \subsection{Results}
    \label{subsec:open_set_results}
    \mypar{Office31} \Cref{tab:office31} compares our method with state-of-the-art UDA and SFDA methods on the Office31 dataset under the open-set setting. For the UDA setting, even though our method does not use source data at all during adaptation, it achieves comparable performance with other methods. Note that the availability of source data in the UDA setting is a considerable advantage, given that they can be used for aligning domains. For the more challenging SF-UDA setting, we achieve the highest HOS score by a notable margin of $+2.1\%$, $+5.0\%$  on recent baselines, such as USD \cite{usd} and AaD \cite{aad}. In addition, our method achieves the highest $\text{HOS}$ score in four source-target configurations, demonstrating to obtain a better balance between $\text{OS}*$ and $\text{UNK}$. On the contrary, previous methods perform well on one but highly degrade on the other. For example, SHOT often achieves the best result on $\text{OS}*$, but performs poorly on $\text{UNK}$, yielding a lower $\text{HOS}$ score. \\
    \\
    \mypar{Office-Home} \Cref{tab:officehome} shows the performance of our method w.r.t. state-of-the-art UDA and SF-UDA methods on Office-Home under the open-set setting. Although our method does not use source data during the adaptation, we outperform several UDA methods, such as STA \cite{open3}, OSBP \cite{open4}, DAOD \cite{daod}, and DANCE \cite{self_supervised7}. In addition, our method surpasses the more recent source-free approaches such as AaD \cite{aad} and USD \cite{usd} with $+2.4\%$ and $+0.1\%$ in $\text{HOS}$, and $+2.6\%$ and $+3.0\%$ in $\text{UNK}$ comparison, demonstrating the robustness of our approach across the board.

    \begin{table}[t!]
    \caption{Ablation study of the components of the proposed method measured by HOS score (\%) on three sub tasks of Office31. First row excludes all the components. Second row includes the clustering initialisation \cref{subsec:initial_clustering}. Third row and fourth row include the two strategies in \cref{subsec:entropy_estimation} for estimating the uncertainty of pseudo-labels, respectively. Last row introduces the proposed NL-InfoNCELoss \cref{subsec:temporal_queue_exclusion}.}
    \begin{adjustbox}{width=\columnwidth}
    \scriptsize
    \begin{tabular}{cccc|c}
    \toprule
        
    \shortstack{Clustering init.} & \shortstack{Neighbours Consensus \\ Uncertainty Estimation} & \shortstack{Class Separation \\ Uncertainty Estimation} & \shortstack{NL-InfoNCELoss} & \shortstack{Avg. \\ Acc.}
    \\ \midrule
    \xmark & \xmark   &  \xmark  &  \xmark   & 35.3
    \\
    \cmark & \xmark   &  \xmark  &  \xmark   & 77.9
    \\
    \cmark & \cmark &  \xmark  &  \xmark   & 80.6
    \\
    \cmark & \cmark & \cmark &  \xmark  & 82.4
    \\
    \cmark & \cmark & \cmark & \cmark & 85.4
    \\ \bottomrule
    \end{tabular}
    \end{adjustbox}
    
    \label{tab:ablation}
    \end{table}
    
    \subsection{Analysis}
    \label{subsec:analysis}

    \mypar{Ablation study} Our method comprises several components to effectively adapt to a target domain in the context of source-free adaptation. \Cref{tab:ablation} shows the benefit of these components: clustering initialisation (\textit{c.f.} \cref{subsec:initial_clustering}), neighbours consensus and class separation uncertainty estimations (\textit{c.f.} \cref{subsec:entropy_estimation}), and the proposed NL-InfoNCELoss (\textit{c.f.} \cref{subsec:temporal_queue_exclusion}). Clustering initialisation brings a major benefit to the training performance, doubling the average accuracy. The benefit in performance of estimating uncertainties is approx. $+2\%$ each. The best performance is achieved when also including the NL-InfoNCELoss, with a performance gain of approx. $+3\%$.\\

\begin{figure}[t]
        \centering
        \includegraphics[width=\linewidth]{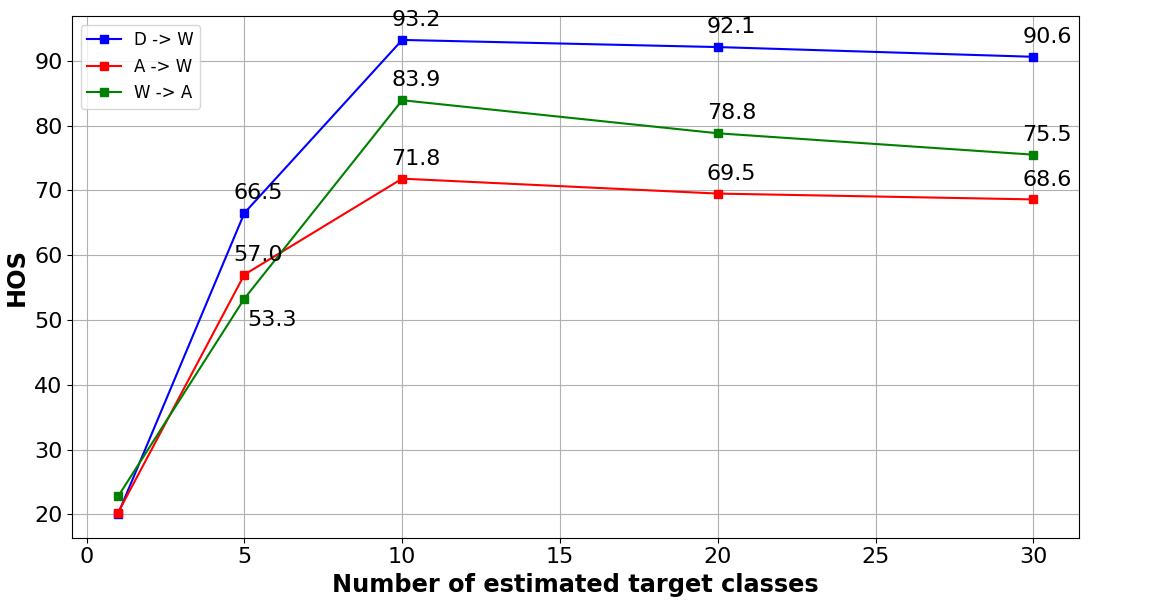}
        \caption{$HOS$ score on three sub tasks of Office31 ($\%$) with a various number of estimated target classes $|C_P|$. The best performance is obtained with $|C_P| = |C_S|$  target classes.}
        \label{fig:num_classes}
    \end{figure}

    \mypar{Cardinality of target-private classes} As detailed in \cref{subsec:initial_clustering}, the number of the target-private classes $|C_P|$ is unknown during adaptation. We set the expected number of target-private $|\widehat{C}_P|$ arbitrarily and we use this value to both extend the size of the classifier output and perform the initial clustering. Here, we study the performance of our method with different values of $|\widehat{C}_P|$. Results of this experimentation are shown in \cref{fig:num_classes}. Particularly, 
    the best performance is obtained when the number of target private classes is equal to the cardinality of source classes, \textit{i.e.} $|C_P| = |C_S|$. Nonetheless, using even higher values for $|C_P|$, our method achieves good performance obtaining an $\text{HOS}$ score close to the one achieved with the optimal value of $C_P$. Contrarily, the worst performance is obtained using $C_P=1$. This results strengthen our hypothesis that predicting all the samples from target-private classes in a unique unknown class is suboptimal, since it requires aggregating semantically different samples together. Following this, we design our method to segregate these samples into multiple unknown classes.\\

    \mypar{From uncertainty to selection probabilities}
    In \cref{eq:weights,eq:weights2}, the uncertainty coefficient $w^{nc}_t$ and $w^{cs}_t$ are estimated with the use of a monotonically decreasing function $\mathcal{F}$. We made two choices for this function: (i) a linear function $\mathcal{F}_{\operatorname{lin}}(x) = 1 - x$; and (ii) an exponential function $\mathcal{F}_{\operatorname{exp}}(x) = e^{-x} $.
    In \Cref{tab:lin_vs_exponential}, we study the effect of using either or both functions to calculate the uncertainty coefficients. Our model achieves comparable performance with all the combinations of these functions.
    As reported in \cite{Litrico}, the exponential function works better for \cref{eq:weights} because it avoids to penalise samples near the boundaries.
    For \cref{eq:weights2}, the effect of the choice of $\mathcal{F}$ is marginal, as the performance drop is approx. $0.1\%$. A motivation of this behaviour is that the minimum value for the coefficient $u^{cs}_t$ is $0.5$, which avoids over-penalising samples when $\mathcal{F}_{\operatorname{lin}}$ is used.\\
    
    \mypar{Combining uncertainty estimation strategies}
    In \cref{subsec:entropy_estimation}, we estimate the uncertainty coefficients $w^{nc}_t$ and $w^{cs}_t$, which are used as the success probability to sample values from Bernoulli distributions. Sampled values are combined together via a logical operator $\oplus$ (\cref{eq:merging_uncertainty}). We opted for two choices for the $\oplus$ operator: \textsc{and} and \textsc{or} operators. We compare the benefits of these operators in \Cref{tab:and_vs_or}.     
    Although either operator surpasses the state-of-the-art, our model achieves the best performance by combining the uncertainty probabilities with the \textsc{and} operator. The \textsc{and} operator requires that samples have low uncertainty in both measures, resulting in a more fine-grained sample selection with respect to the \textsc{or} operator.

    \begin{table}[t!]
    \caption{HOS score (\%) on three sub-tasks of Office31 using different choices of the function $\mathcal{F}$ in \cref{eq:weights,eq:weights2}.}
    \begin{adjustbox}{width=\columnwidth}
    \scriptsize
    \begin{tabular}{cc|ccc|c}
    \toprule
        
    \cref{eq:weights} & \cref{eq:weights2} & D $\rightarrow$ W & A $\rightarrow$ W & D $\rightarrow$ A & \shortstack{Avg.}
    \\ \midrule
    $\mathcal{F}_{\operatorname{lin}}$ & $\mathcal{F}_{\operatorname{lin}}$ & 91.5 & 79.0 & 76.2 & 82.2
    \\
    $\mathcal{F}_{\operatorname{lin}}$ & $\mathcal{F}_{\operatorname{exp}}$ & 90.5 & 77.4 & 77.8 & 81.9
    \\
    $\mathcal{F}_{\operatorname{exp}}$ & $\mathcal{F}_{\operatorname{exp}}$ & 93.1 & 83.6 & 78.0 & 84.9
    \\    \midrule
    $\mathcal{F}_{\operatorname{exp}}$ & $\mathcal{F}_{\operatorname{lin}}$ & 93.2 & 83.9 & 78.0 & \textbf{85.0}
    \\ \bottomrule
    \end{tabular}
    \end{adjustbox}
    
    \label{tab:lin_vs_exponential}
    \end{table}

\begin{table}[t]
        \caption{$HOS$ score on three sub tasks of Office31 ($\%$) comparing different choices of the logical operator $\oplus$ in \cref{eq:merging_uncertainty}.}
        \centering
        \begin{adjustbox}{width=\columnwidth}
        \begin{tabular}{l|ccc|c}
            \toprule
             Method & D $\rightarrow$ W & A $\rightarrow$ W & D $\rightarrow$ A & Avg.\\ \midrule
             Ours w/ OR sampling & 91.5 & 80.1 & 77.7 & 83.1\\
             Ours w/ AND sampling & 93.2 & 83.9 & 78.0 & 85.0\\ 
            \bottomrule     
        \end{tabular}
        \end{adjustbox}
        \label{tab:and_vs_or}
    \end{table}

    \begin{table}[t]
        \caption{Clustering accuracy ($\%$) on three sub tasks of Office31 evaluating the ability of our model to perform novel class discovery. Note that accuracy is calculated only for samples belonging to target-private classes.}
        \centering
        \begin{adjustbox}{width=0.6\columnwidth}
        \begin{tabular}{ccc|c}
            \toprule
             D $\rightarrow$ W & A $\rightarrow$ W & D $\rightarrow$ A & Avg.\\ \midrule
             71.3 & 54.8 & 66.7 & 64.2\\
            \bottomrule     
        \end{tabular}
        \end{adjustbox}
        \label{tab:private_acc}
    \end{table}

    \mypar{Discovery of the underlying semantics of novel classes}
    We present here additional experiments to asses whether our model is able, as a byproduct, to aggregate samples from target-private classes into different clusters based on the semantics. To achieve this objective, the model has to learn the semantics of target-private classes, even if it cannot be directly optimised for this scope, due to the lack of labels in the target domain.
    We follow the same protocol as in state-of-the-art novel class discovery works \cite{ncd1, ncd2}.
    We train the model by setting $|C_P|$ as the actual number of private classes: this is necessary to evaluate the results of these experiments.
    Since we do not use labels for target-private classes during adaptation, a class misalignment is present between the predicted and the ground-truth target-private classes. To solve this problem, at inference time, we match the predicted target-private classes against the ground-truth classes. Specifically, we compute target-private class prototypes by averaging features of samples belonging to each class, according to their ground truth labels. The same process is repeated to compute the estimated prototypes using predictions rather than labels. We then run the Hungarian algorithm to perform the matching between the ground-truth and the estimated prototypes, and adjust the predicted labels accordingly. 
    Once the predicted and ground-truth target-private classes are aligned, we quantitatively and qualitatively assess the ability of our model to correctly classify samples from target-private classes.
    As in previous works\cite{ncd1, ncd2}, we quantitatively evaluate the ability of our method to cluster samples from target-private classes by computing the clustering accuracy:

    \begin{equation}
        \operatorname{ClusterAcc} = \max_{t\in \mathcal{P}(C^P)} \frac{1}{|C^P|} \sum_{j=1}^{|C^P|} \mathds{1}(y_j = t(\hat{y}_j))
    \end{equation}
    where $y_j$ and $\hat{y}_j$ are ground-truth labels and predictions, respectively. For each target-private sample, $\mathcal{P}(C^P)$ is the set of all possible permutations of $C^P$ classes and $t$ is a permutation.
    
\begin{figure}
    \centering
    \includegraphics[width=0.8\linewidth]{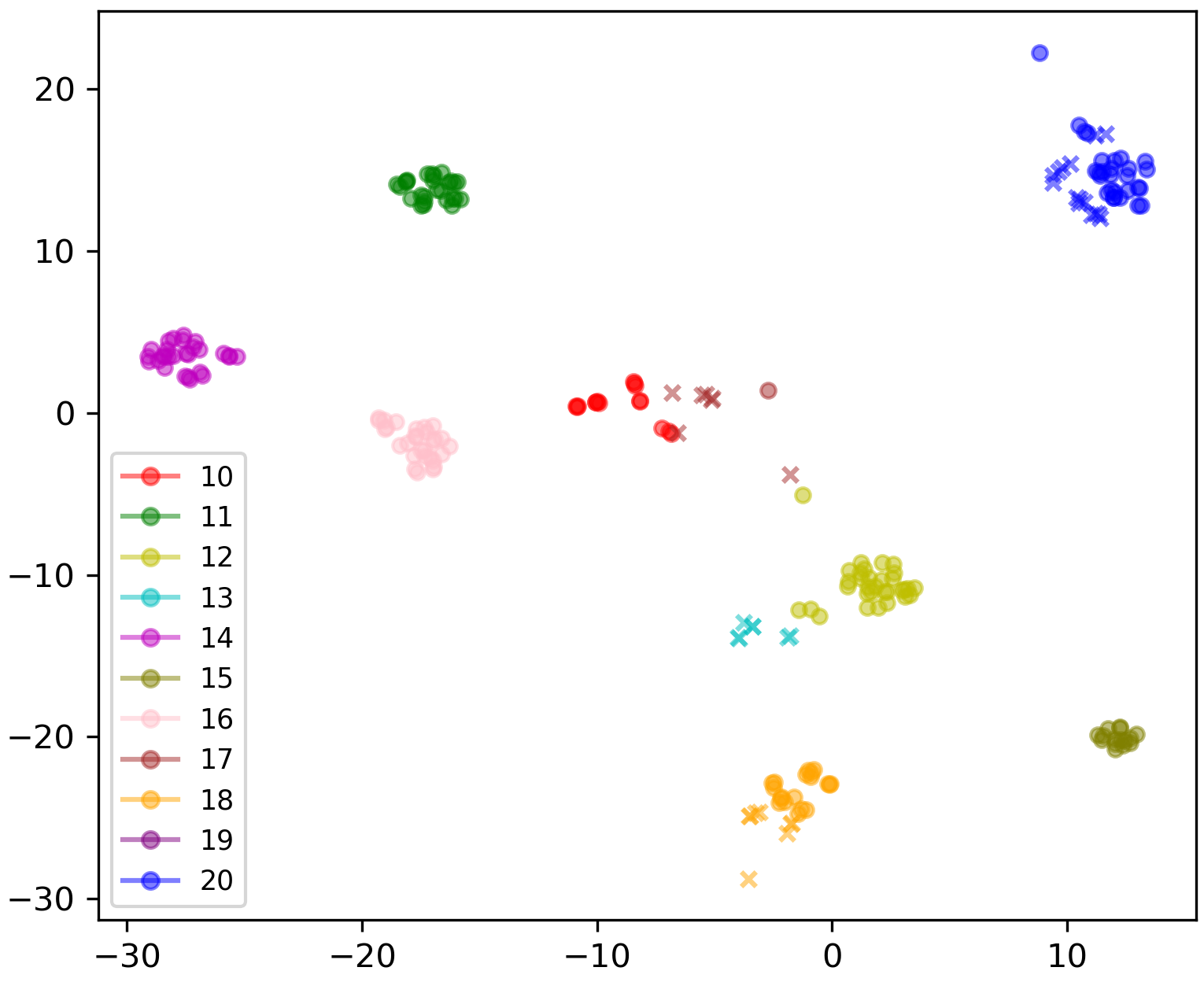}
    \caption{Feature space visualization for target-private classes. The colour of samples represents the predicted class. The $\bullet$ symbol represents correct predictions, whereas the $\times$ represents incorrect predictions. Our model produces a well-structured features space for target-private classes, where samples belonging to the same novel classes are clustered together. This opens the possibility of performing novel class discovery.}
    \label{fig:private_acc}
    \end{figure}

    \Cref{tab:private_acc} and \Cref{fig:private_acc} show that our model is able to achieve satisfying results in the classification of target-private samples, even if it has been neither optimised for this task nor trained with labels from these classes. This means the model has learned the underlying semantics of novel classes, enabling the possibility to potentially perform novel category discovery.
    Note that this behaviour is not possible with previous works that classify all the private samples in a single unknown class.
    Finally, \Cref{fig:private_acc} shows the ability of our model to produce a well-structured features space, where samples belonging to the same target-private class lie close in the space, allowing them to be clustered in the same novel class.

    \section{Conclusion}
    \label{sec:conclusion}
    In this paper, we introduced a novel approach to address the challenges of Source-Free Open-Set Domain Adaptation (SF-OSDA) for image classification. Unlike existing approaches that group samples from unknown classes into a single ``unknown'' class, our method leverages the granularity of target-private classes to segregate their samples into multiple clusters, resulting in a more effective adaptation that enables the learning of the underlying semantics of novel classes. To that end, we introduce a novel uncertainty estimation technique to handle noisy pseudo-labels produced during the adaptation. We also propose a novel \textit{NL-InfoNCELoss} that integrates  negative learning into self-supervised contrastive learning further increasing the robustness of the contrastive framework to the noise in the pseudo-labels. Our approach outperforms state-of-the-art methods on benchmark datasets, such as Office31 and Office-Home. Further analyses show that our model can aggregate samples from target-private classes into different clusters as a byproduct of the adaptation, as it has learned the underlying semantics of novel classes. This behaviour also opens the possibility of performing novel class discovery.


\bibliography{bibtex}
\bibliographystyle{IEEEtran}

\begin{IEEEbiography}[{\includegraphics[width=1in,height=1.25in,clip,keepaspectratio]{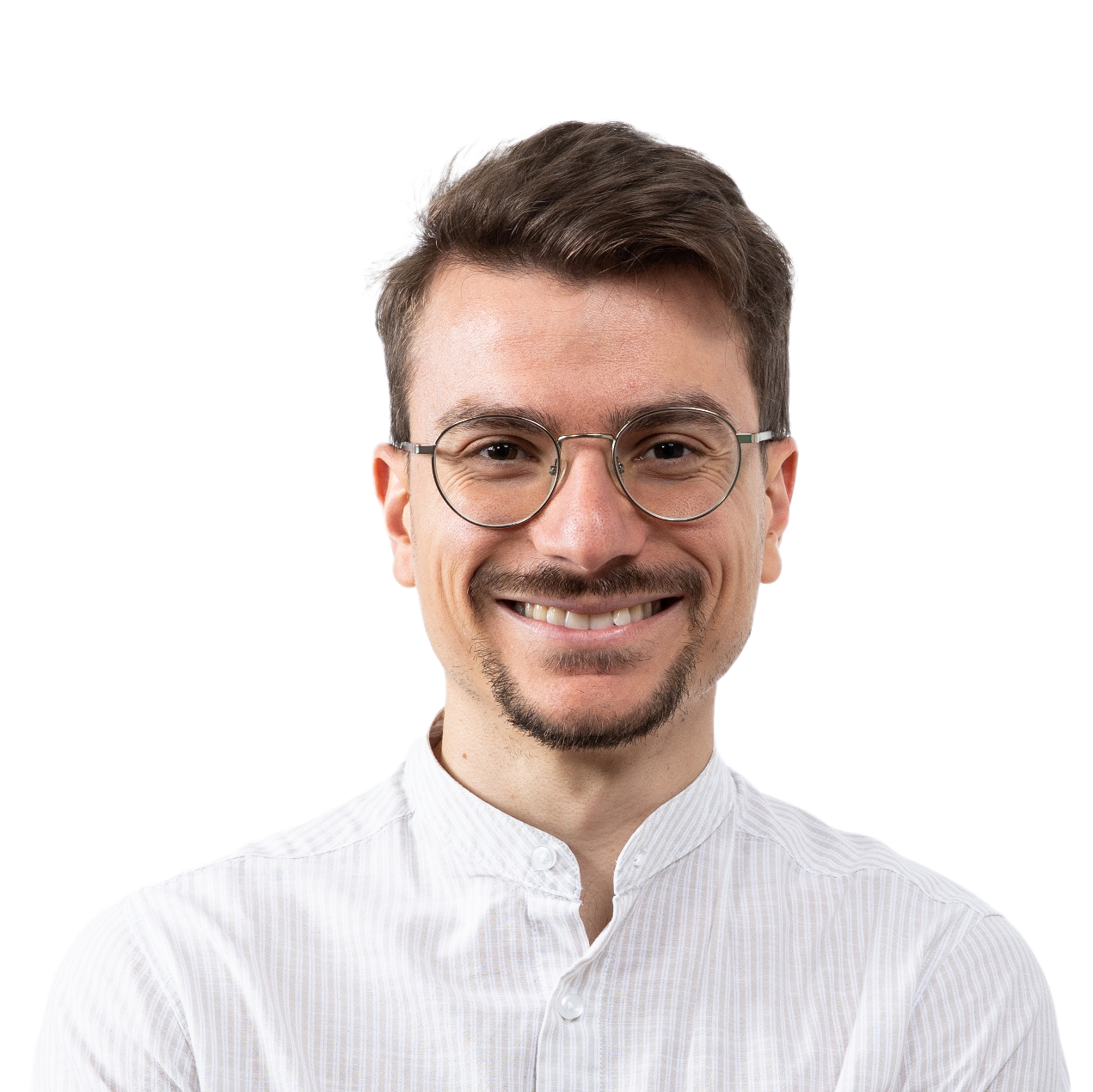}}]{Mattia Litrico}
 Mattia Litrico is a Ph.D. Student in Deep Learning and Computer Vision at the University of Catania. He obtained his M. Sc. in Computer Science at the University of Catania (Italy) in 2021. He was a Research Fellow at the Italian Institute of Technology (IIT), Pattern Analysis and Computer Vision (PAVIS) department in 2022. His research interests include machine learning and computer vision, with a particular emphasis on domain adaptation techniques. 
\end{IEEEbiography}

\begin{IEEEbiography}[{\includegraphics[width=1in,height=1.25in,clip,keepaspectratio]{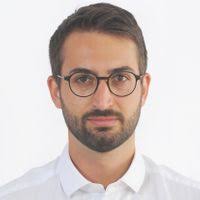}}]{Davide Talon}
 Davide Talon is a Postdoc at Fondazione Bruno Kessler. He obtained his M. Sc. in Computer Engineering at University of Padova (Italy). He pursued a PhD degree in Computer Science at the Italian Institute of Technology (IIT), Pattern Analysis and Computer Vision (PAVIS) department. His research interests span across machine learning, deep learning, representation learning and causality. 
\end{IEEEbiography}

\begin{IEEEbiography}[{\includegraphics[width=1in,height=1.25in,clip,keepaspectratio]{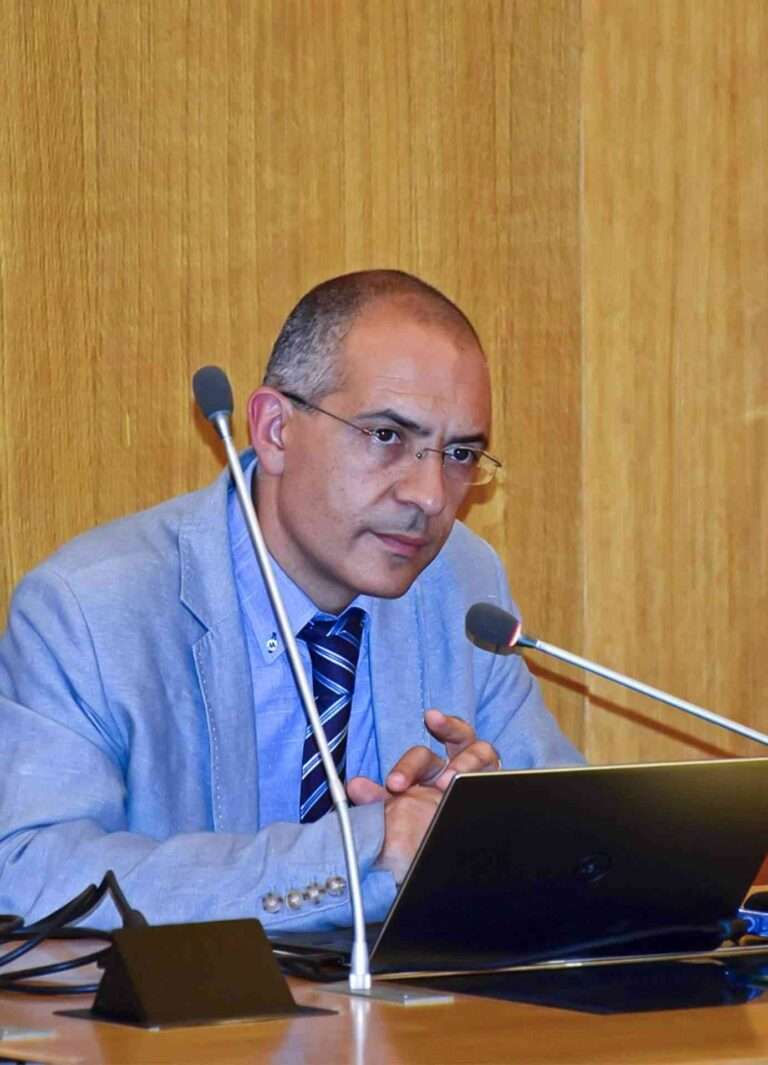}}]{Sebastiano Battiato}
  Sebastiano Battiato (Senior Member, IEEE) is a Full Professor of Computer Science at the University of Catania. Throughout his career, Battiato has held significant leadership roles such as Chairman of the Undergraduate Program in Computer Science, Rector's delegate for postgraduate education, and Scientific Coordinator of the PhD Program in Computer Science. His research interests span Computer Vision, Imaging Technology, and Multimedia Forensics, focusing on enhancing imaging algorithms for both acquisition and post-processing. Battiato is a prolific scholar, having edited six books and authored around 350 papers. He has been a principal investigator in numerous research projects, supervised many PhD students and postdocs, and is a co-inventor of 25 international patents. His contributions to computer science have been recognized with several awards, including the 2017 PAMI Mark Everingham Prize and the 2011 Best Associate Editor Award of the IEEE Transactions on Circuits and Systems for Video Technology. 
\end{IEEEbiography}

\begin{IEEEbiography}[{\includegraphics[width=1in,height=1.25in,clip,keepaspectratio]{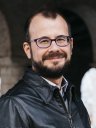}}]{Alessio Del Bue} Alessio Del Bue
 (Member, IEEE) is a Tenured Senior Researcher leading the Pattern Analyisis and computer VISion (PAVIS) Research Line of the Italian Institute of Technology (IIT), Genoa, Italy. He is a coauthor of more than 100 scientific publications in refereed journals and international conferences on computer vision and machine learning topics. His current research interests include 3D scene understanding from multi-modal input (images, depth, and audio) to support the development of assistive artificial intelligence systems. He is a member of the technical committees of major computer vision conferences (CVPR, ICCV, ECCV, and BMVC). He serves as an Associate Editor for Pattern Recognition and Computer Vision and Image Understanding journals. He is a member of ELLIS.
\end{IEEEbiography}

\begin{IEEEbiography}[{\includegraphics[width=1in,height=1.25in,clip,keepaspectratio]{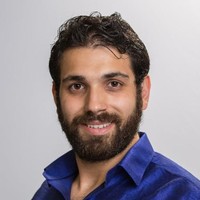}}]{Mario Valerio Giuffrida}
 Mario Valerio Giuffrida is an Assistant Professor in Computer Vision at The University of Nottingham. He pursued a PhD degree in Image Analysis at IMT School For Advanced Studies Lucca. In the 2016, he participated at the enrichment program of The Alan Turing Institute. His current research interests include machine learning and computer vision for plant phenotyping.
\end{IEEEbiography}

\begin{IEEEbiography}[{\includegraphics[width=1in,height=1.25in,clip]{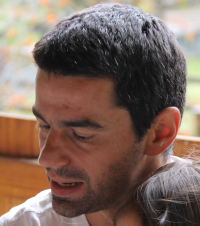}}]{Pietro Morerio} Pietro Morerio received his M. Sc. in Theoretical Physics from the University of Milan (Italy) in 2010 (summa cum laude). He was Research Fellow at the University of Genoa (Italy) from 2011 to 2012, working in Video Analysis for Interactive Cognitive Environments. He pursued a PhD degree in Computational Intelligence at the same institution in 2016. Currently he is a Researcher at Istituto Italiano di Tecnologia (IIT), Pattern Analysis and Computer Vision (PAVIS) department, his research including machine learning, deep learning and computer vision, with a particular focus on multimodal learning.
\end{IEEEbiography}
\vfill

\end{document}